\documentclass{article}

\usepackage{microtype}
\usepackage{graphicx}
\usepackage{booktabs}
\usepackage{hyperref}
\usepackage{amsmath}
\usepackage{amssymb}
\usepackage{amsthm}
\usepackage[accepted]{icml2024}
\usepackage[labelfont=it,font=small]{caption}
\usepackage{subcaption}
\usepackage{mathtools}
\usepackage{csquotes}
\usepackage{makecell}
\usepackage{multirow}
\usepackage[capitalize,noabbrev]{cleveref}

\theoremstyle{plain}
\newtheorem{theorem}{Theorem}[section]
\newtheorem{proposition}[theorem]{Proposition}

\theoremstyle{definition}
\newtheorem{definition}[theorem]{Definition}
\theoremstyle{remark}
\newtheorem{remark}[theorem]{Remark}

\usepackage{pifont}
\newcommand{\cmark}{\ding{51}}
\newcommand{\xmark}{\ding{55}}

\begin{document}

\twocolumn[
\icmltitle{Neural Operators with Localized Integral and Differential Kernels}
\icmlsetsymbol{equal}{*}

\begin{icmlauthorlist}
\icmlauthor{Miguel Liu-Schiaffini}{equal,calt}
\icmlauthor{Julius Berner}{equal,calt}
\icmlauthor{Boris Bonev}{equal,nvda}
\icmlauthor{Thorsten Kurth}{nvda}
\icmlauthor{Kamyar Azizzadenesheli}{nvda}
\icmlauthor{Anima Anandkumar}{calt}
\end{icmlauthorlist}

\icmlaffiliation{calt}{Department of Computing and Mathematical Sciences, California Institute of Technology, Pasadena CA 91125}
\icmlaffiliation{nvda}{NVIDIA, Santa Clara, CA 95051}

\icmlcorrespondingauthor{Miguel Liu-Schiaffini}{\href{mailto:mliuschi@caltech.edu}{mliuschi@caltech.edu}}
\icmlcorrespondingauthor{Julius Berner}{\href{mailto:jberner@caltech.edu}{jberner@caltech.edu}}
\icmlcorrespondingauthor{Boris Bonev}{\href{mailto:bbonev@nvidia.com}{bbonev@nvidia.com}}
\icmlkeywords{Neural Operators, Scientific Machine Learning, Convolutions}
\vskip 0.3in
]

\printAffiliationsAndNotice{\icmlEqualContribution} 

\begin{abstract}
Neural operators learn mappings between function spaces, which is practical for learning solution operators of PDEs and other scientific modeling applications. Among them, the Fourier neural operator (FNO) is a popular architecture that performs global convolutions in the Fourier space. However, such global operations are often prone to over-smoothing and may fail to capture local details. In contrast, convolutional neural networks (CNN) can capture local features but are limited to training and inference at a single resolution. In this work, we present a principled approach to operator learning that can capture local features under two frameworks by learning differential operators and integral operators with locally supported kernels. Specifically, inspired by stencil methods, we prove that we obtain differential operators under an appropriate scaling of the kernel values of CNNs. To obtain local integral operators, we utilize suitable basis representations for the kernels based on discrete-continuous convolutions. Both these approaches preserve the properties of operator learning and, hence, the ability to predict at any resolution. Adding our layers to FNOs significantly improves their performance, reducing the relative $L^2$-error by $34$-$72$\% in our experiments, which include a turbulent 2D Navier-Stokes and the spherical shallow water equations.
\end{abstract}

\section{Introduction}
\label{sec:introduction}

\begin{figure}[t!]
    \begin{subfigure}{\linewidth}
        \centering
        \includegraphics[width=\linewidth]{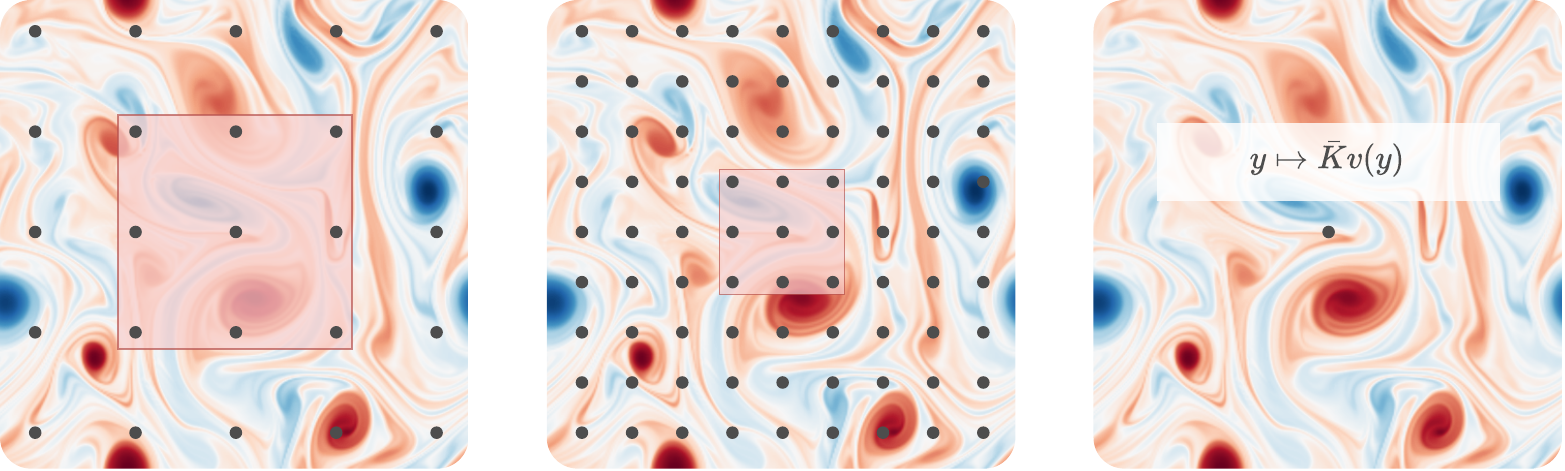}%
        % \vspace{0.15em}
        \caption*{Regular kernel}
        \vspace{0.45em}
    \end{subfigure}
    \begin{subfigure}{\linewidth}
        \centering
        \includegraphics[width=\linewidth]{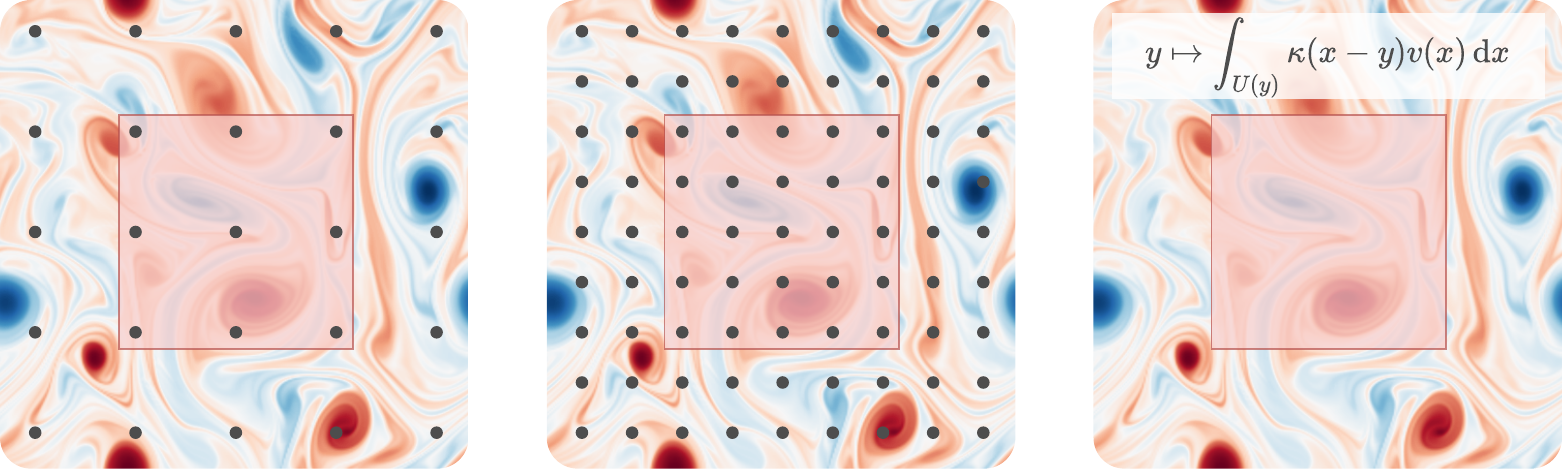}%
        % \vspace{0.15em}
        \caption*{Local integral kernel}
        \vspace{0.45em}
    \end{subfigure}
    %\vspace{0.3em}
    \begin{subfigure}{\linewidth}
        \centering
        \includegraphics[width=\linewidth]{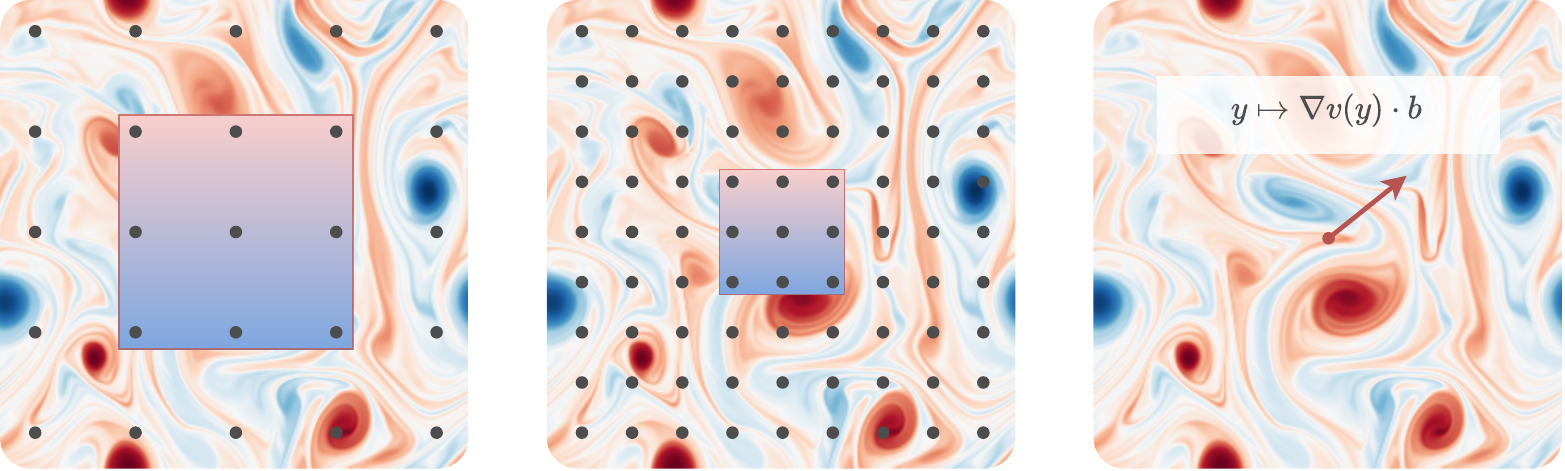}%
        % \vspace{0.15em}
        \caption*{Differential kernel}
        %\vspace{-0.25em}
    \end{subfigure}
    \caption*{$ h \hspace{2.2em} \xrightarrow{\mathmakebox[3em]{}} \hspace{2.2em} h/2 \hspace{2.2em}\xrightarrow{\mathmakebox[3em]{}} \hspace{2.2em} 0$}
    \vspace{0.4em}%
    \caption{Visualization of different limits of a convolution with a discretized function $v$ as the grid width $h$ is refined, i.e., $h\to 0$. \textbf{(Top)} A regular convolution is collapsing to a pointwise linear operator. \textbf{(Middle)} Instead, we can use a kernel that can be evaluated at arbitrary resolutions and keep the receptive field unchanged, to converge to a local integral operator, see \Cref{sec:disco}. \textbf{(Bottom)} Alternatively, we can let it collapse while constraining the kernel appropriately, converging to a differential operator, see \Cref{sec:diff_layer}.}
    \label{fig:diffconv}
    \vspace{-0.5em}
\end{figure}

Deep learning holds the promise to greatly accelerate advances in computational science and engineering, which often require numerical solutions of partial differential equations (PDEs)~\cite{azizzadenesheli2024neural,zhang2023artificial,cuomo2022scientific}. Recent advances in deep learning have enabled applications such as weather forecasting~\citep{Pathak2022,Bonev2023,lam2023learning}, seismology~\citep{sun2023phase,shi2023broadband}, reservoir engineering for carbon capture~\citep{wen2022u,wen2023real}, and many other applications with orders of magnitude speedup over traditional methods. 

Many of the above results are achieved by neural operators, which learn mappings between function spaces, enabling operator learning for function-valued data~\citep{li2021physics,azizzadenesheli2024neural,raonic2023convolutional}. In particular, they are agnostic to the discretization of the input and output functions---a vital property in the context of PDEs where data is often provided at varying resolutions and high-resolution data is costly to generate~\cite{Kovachki2021}. In contrast, standard neural networks such as convolutional neural networks (CNN)~\citep{ronneberger2015u,gupta2022towards} require the functions to be discretized at a fixed resolution on a regular grid, which is limiting.

In the past few years, various architectures of neural operators have been developed. Among them, the Fourier neural operator (FNO)~\cite{Li2020}, which performs global convolutions in Fourier space, has gained popularity and shown good performance in a number of applications. However, such global operations are often prone to over-smoothing and may fail to capture local details. Several other architectures instead learn global operators in physical space (e.g., \citep{li2022transformer}), but they likewise lack the inductive bias of local operations. 

There are many applications that require a local neural operator. For instance, solution operators of several relevant PDEs are of local nature. Examples include hyperbolic PDEs, which have real-valued characteristic curves \cite{LeVeque1992}. As a result, a solution at a given point will only depend on the initial condition within a neighborhood of that point. As such, their solution operators only have a local receptive field and can, therefore, be efficiently learned by locally supported kernels.

Further examples of local operators are differential operators, which can be expressed in terms of pointwise multiplication with the frequency in the spectral domain. Consequently, they introduce large errors when approximated by a finite number of parameters in Fourier space.
In this context, we note that the emulation of classical numerical methods to solve partial differential equations, such as finite-difference methods, relies on the usage of local stencils for differentiation~\cite{thomas2013numerical}. This calls for the presence of local and differential operators in neural operator architectures.

\begin{table}[t]
\vskip 3pt
\caption{Comparison of different architectures for the solution of PDEs. The top half enumerates architectures for planar domains and the bottom half for spherical domains. Our proposed architectures are highlighted in bold and the differential and integral kernels are detailed in~\Cref{sec:local_convs}.}
\label{tab:compare}
\vspace{-0.5em}
\begin{center}
\begin{tabular}{l@{\hspace{-0.1cm}}c@{\hspace{0.1cm}}c@{\hspace{0.1cm}}c}\toprule
\multirow{2}*{Architecture} & \multirow{2}*{Efficient}  & Receptive & no input \\ 
&  & field & downsampling\\ \midrule
GNO & \xmark & local/global & \cmark  \\
FNO & \cmark & global & \cmark  \\
CNO / U-Net & \cmark & local & \xmark  \\
\textbf{FNO + integral} & \cmark & local/global & \cmark \\
\textbf{FNO + differential} & \cmark & local/global & \cmark  \\ \midrule
SFNO & \cmark & global & \cmark \\
\textbf{SFNO + integral} & \cmark & local/global & \cmark \\
 \bottomrule
\end{tabular}
\end{center}
\vspace{-0.5em}
\end{table}

Naturally, every local operation can also be represented by a global operation. However, this is typically vastly parameter-inefficient and does not provide a good inductive bias for learning local operations. In the context of neural operators, spectral variants of neural operators, such as the FNO~\citep{Li2020} and Spherical FNO (SFNO)~\citep{Bonev2023}, are theoretically able to approximate local convolutions. However, representing local kernels requires the approximation of a global signal in the spectral domain, in turn demanding a large number of parameters (due to the uncertainty principle \cite{Cohen-Tannoudji1977}).

A few prior works have explored the usage of local operations in the context of neural operators. \citet{li2020neural} introduce graph neural operators (GNOs), which parameterize local integral kernels with a neural network. However, evaluating this network on all combinations of points to compute the integrals can make GNOs computationally expensive and slower than hardware-optimized convolutional kernels. Alternatively, \citet{ye2022learning,ye2023locality} propose local neural operators by combining FNOs with convolutional layers and, similarly,~\citet{wen2022u} integrate U-Nets and FNOs. Moreover,~\citet{raonic2023convolutional} propose convolutional neural operators (CNOs) by leveraging U-Nets that (approximately) respect the bandlimits. 

However, since all of the above approaches rely on standard convolutional layers on equidistant grids, they have the following shortcomings.
First, such approaches do not allow for a natural extension to unstructured grids or other geometries, which are ubiquitous in PDE problems~\cite{li2023geometry}.
Moreover, they can only be applied to higher resolutions by downsampling of the (intermediate) inputs to the training resolution. Important high-frequency content can be lost through downsampling, which is particularly problematic for multi-scale data in the context of PDEs. 
In contrast, we develop convolutional layers that can be applied at any resolution without downsampling. We achieve this by appropriately scaling the receptive field or the values of the kernel (see \Cref{fig:diffconv}).

To summarize, current neural operator architectures suffer from at least one of the following limitations (see \Cref{tab:compare}): (1)~they cannot succinctly represent operations with a local receptive field, e.g., FNO, or (2)~they cannot be applied to different resolutions without relying on explicit up-/downsampling which may degrade performance, e.g., CNO, or (3)~they cannot be scaled to obtain sufficient expressivity, since they incur prohibitively high computationally costs, such as, for instance, GNOs. 

\paragraph{Our approach:} In this work, we develop 
computationally efficient and principled approaches to include operations in neural operators that capture local receptive fields while retaining the ability to approximate operators and, hence, extend to multiple resolutions.
We consider two kinds of localized operators: differential operators and integral kernel operators with a locally supported kernel (see~\Cref{fig:diffconv}). 

For the first, we draw inspiration from stencils of finite-difference methods. We derive conditions to modify convolutional layers such that they converge to a unique differential operator when the discretization is refined. For the second case of local integral operators, we adapt discrete-continuous (DISCO) convolutions~\cite{Ocampo2022} to provide an efficient, discretization-agnostic framework that can be applied to general meshes on both planar and spherical geometries. To our knowledge, we are the first to connect the DISCO framework to operator learning.

\begin{figure}[tb]
    \centering
    \includegraphics[width=0.85\linewidth]{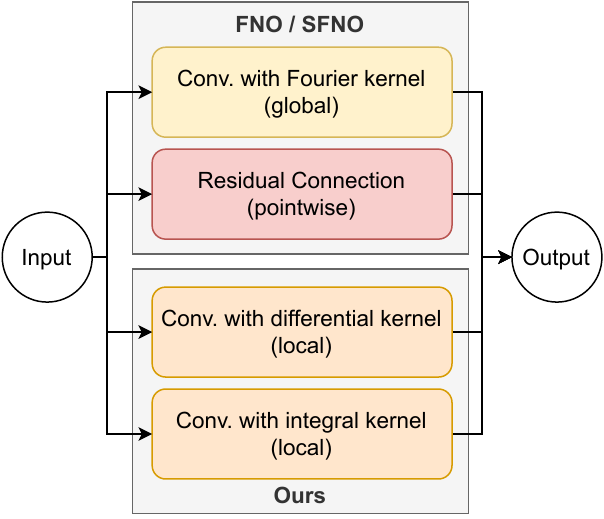}
    \caption{A single layer of our local neural operator. We add (up to) two local operations using the convolutions with differential kernel (\Cref{sec:diff_layer}) and local integral kernel (\Cref{sec:disco}).}
    \label{fig:layer}
    \vspace*{-0.5em}
\end{figure}

Finally, we devise efficient implementations of both layers and show that the inductive bias of local operations (see \Cref{fig:layer}) can significantly improve the performance of FNOs and SFNOs on three different benchmarks. In particular, we learn a differential operator in a Darcy flow setting for motivating the need for neural operators with local inductive biases, and we improve over FNO by $87\%$. Additionally, we improve over the baseline (S)FNO by $34\%$ on turbulent 2D Navier-Stokes equations, by $72\%$ on the shallow water equations on the sphere, and by $63\%$ on the 2D diffusion-reaction equation. We also apply our method on modeling a flow past a cylinder, discretized on an irregular grid. We improve upon the best baseline by $42\%$.

\paragraph{Outline:} The remainder of the paper is organized as follows: Section \ref{sec:related_work} outlines connections to other architectures and ideas within the neural operator literature. Section \ref{sec:local_convs} introduces the main ideas of the paper as well as the macro-architecture of our proposed neural operator. Section \ref{sec:experiments} discusses numerical experiments and results, and Section \ref{sec:conclusion} summarizes the findings of our work.
\section{Related work and connections to other frameworks}
\label{sec:related_work}

We strive to formulate local operations that are consistent with the neural operator paradigm, which stipulates that operations within the model are independent of the discretization of input functions~\cite{Kovachki2021}. In principle, this allows the evaluation of the neural operator on arbitrary meshes\footnote{As long as they are suitable for the evaluation of the numerical operations of the neural operator, such as the Discrete Fourier Transform used in the Fourier neural operator (see Section~\ref{sec:local_convs}).}.
Several discretization-independent local operations have been introduced in the deep learning literature. We outline the connections between these works in the following.
\paragraph{GNOs and Hypernetworks:} In general, local operations can be provided by graph neural operators (GNOs)~\cite{li2020neural}. However, in their general form, GNOs are typically slow, hard-to-train, and not equivariant w.r.t.\@ the symmetry group of the underlying domain~\cite{li2020multipole}. Taking a convolutional kernel, GNOs subsume a series of works on hypernetwork approaches for convolutional layers~\cite{wang2018deep,shocher2020discrete}.

\paragraph{DISCO Convolutions:} We show that DISCO convolutions~\cite{Ocampo2022} represent special cases of GNOs with a convolutional kernel, which enables generalization to different geometries and an efficient implementation, since the kernel can be pre-computed (see~\Cref{sec:local_convs}). In particular, DISCO convolutions are typically implemented as learnable linear combinations of fixed basis functions instead of neural networks. Also, they define maps that are equivariant up to the error resulting from the quadrature rule.

\paragraph{Scale-Equivariant CNNs:}
In the area of computer vision, there has been a series of adaptations of CNNs to be (locally) scale-equivariant by using filter dilation, filter rescalings in the discrete domain~\cite{rahman2023truly,sosnovik2021disco, worrall2019deep} or the continuous domain~\cite{xu2014scale,sosnovik2019scale,ghosh2019scale}, or input rescalings~\cite{marcos2018scale}. 
Moreover, filter rescaling has also been explored for more general group-convolutions~\cite{bekkers2019b}.

\paragraph{Neural operators and convolutional layers:} 

While neural operators have been developed independently of the above approaches, one can leverage similar ideas. In particular, one can rescale (i.e., up- and downsample) input and output functions of a convolutional layer or the kernel itself to obtain neural operators. We note that these approaches are, in principle, also applicable to a trained network (without the need for retraining). 

In combination with FNOs, interpolation of the input and output functions has been explored by~\citet{wen2022u}. Also,~\citet{ye2023locality,ye2022learning} emphasized the need for local convolutions in FNOs, however, they do not discuss the application to different discretizations. Combined with correct treatment of the bandlimit of the functions (based on~\citet{karras2021alias}), the convolutional neural operator~\cite{raonic2023convolutional} also uses up- and downsampling of the input and output functions to apply a U-Net architecture.

We note that for bandlimited functions sampled above the Nyquist frequency, the discrete convolutions have a unique correspondence to convolutions in function space, representing a special case of the DISCO framework. However, downsampling the input function introduces errors in cases where the function is not bandlimited.
\section{Local Layers}
\label{sec:local_convs}

In this section, we present two conceptually different types of local layers for neural operators. First, we present efficient integral transforms with local kernels. Then, we focus on the approximation of differential operators. We consider an input function 
\begin{equation*}
    v\colon\mathbb{R}^d \supset D \to \mathbb{R}^n
\end{equation*}
that is discretized on meshes $D^h \subset D$ with width $h$ on a domain $D\subset \mathbb{R}^d$. For notational convenience, we present most ideas for the case $d=1$, but it is straightforward to extend them to higher-dimensional domains.

\subsection{Motivation: Convolutional Layer}
\label{sec:standard_conv}
We take inspiration from convolutional layers since they represent the prototypical version of an efficient, local operation in neural networks. However, we will see that they are not consistent in function spaces; in particular, they converge to a pointwise linear operator when we refine the discretization of the input function $v$.

Let us start by recalling the definition of a convolutional\footnote{In line with deep learning frameworks, we consider convolution with the reflected filter, also known as \emph{cross-correlation}.} layer, specifically a stride-$1$ convolution with $n$ input channels, a single\footnote{This is for notational convenience; the extension to multiple output channels is straightforward.} output channel, and kernel $K=(K_i)_{i=1}^S \subset \mathbb{R}^{n}$ of (odd) size $S$. Assuming a regular grid, i.e., $D^h = \{x_j\}_{j=1}^m \subset \mathbb{R}$ with $x_{j+1} - x_j = h$, we can define the output of the convolutional layer at $y \in D^h$ as
\begin{align}
\label{eq:conv}
    \operatorname{Conv}_K[v](y) &= (K \star \{v(x_j)\}_{j=1}^m)(y) \nonumber\\ &= \sum_{i=1}^S K_{i} \cdot v(z_i + y),
\end{align}
with
\begin{equation}
\label{eq:conv_eval}
    z_i=h\left(i - 1 - \tfrac{S-1}{2}\right),
\end{equation}
where we use zero-padding, i.e., $v(x) =0$ for $x \notin D$.

If we now take the same kernel $K$ for finer discretizations, i.e., $h\to 0$, we see from~\eqref{eq:conv_eval} that $z_i \to 0$ and therefore, 
\begin{equation*}
   \lim_{h \to 0} \, \operatorname{Conv}_K[v](y)= \bar{K} \cdot v(y) \quad \text{with} \quad \bar{K} = \sum_{i=1}^S K_{i},
\end{equation*}
given that the function $v$ is continuous at $y$. In other words, the receptive field with respect to the underlying domain $D$ is shrinking to a point, and the convolutional layer is converging to a \emph{pointwise linear operator}. One way to circumvent this issue would be to downsample the function $v$ appropriately to a pre-defined grid. This is done for previous approaches mentioned in~\Cref{sec:related_work}, at the cost of losing high-frequency information in the input.

In the following, we will present two ways of working on different input resolutions, while not collapsing to a pointwise operator, see also Figure~\ref{fig:diffconv}. First, we show that rescaling~\eqref{eq:conv} by the reciprocal resolution $\frac{1}{h}$ and constraining the kernel $K$ leads to \emph{differential operators}. Then, we define the kernel $K$ as the evaluation of a function over a fixed input domain, leading to \emph{integral operators}.

\subsection{Differential Layer}
\label{sec:diff_layer}

In this section, we construct a layer that converges to a differential operator when the width $h$ of the discretization $D^h$ tends to zero. To prevent the operator from collapsing to a pointwise operation, we constrain and rescale the values of the kernel (according to the discretization width). Note that without rescaling, we would again recover a pointwise operator; however, due to the rescaling by a factor of $\frac{1}{h}$, the limit might not exist without constraining the values of the kernel. The next proposition shows that it is sufficient to subtract the average kernel value.  See~\Cref{app:diff} for a more general statement and a corresponding proof.

\begin{proposition}[First-order differential layer]
\label{prop:diff}
Let $D_h\subset \mathbb{R}^d$ be a regular grid of width $h$ and let $v\in C^1(D, \mathbb{R}^n)$. Then, for every kernel $(K_i)_{i=1}^S \subset \mathbb{R}^{n}$, there exists $(b_j)_{j=1}^n\subset\mathbb{R}^d$ such that 
\begin{equation*}
    \lim_{h \to 0} \, \frac{1}{h} \operatorname{Conv}_{K-\bar{K}}[v](y) =   \sum_{j=1}^n \nabla v_j(y) \cdot b_j
\end{equation*} 
for every $y\in D_h$ , where $\bar{K}=\sum_{i=1}^S K_i.$
\end{proposition}

\Cref{prop:diff} shows that we can learn different directional derivatives using an appropriate adaptation of standard convolutional layers as in~\Cref{sec:standard_conv}. Specifically, we center the kernel $K$ by subtracting its mean $\bar{K}$ and scale the result by the reciprocal resolution $\frac{1}{h}$. In~\Cref{app:diff_empirical}, we empirically evaluate the convergence in resolution to a unique differnetial operator on a simple example.

\begin{remark}[Higher-order differential operator]

Similarly, we could approximate $k$-th order differential operator with further constraints on the elements of $K$ and a scaling factor of $\frac{1}{h^k}$. However, we do not implement them in practice since we can also approximate higher-order derivatives by composing first-order differential layers.
\end{remark}

\subsection{Integral Kernel Layers}

Instead of rescaling the kernel, we can also adapt the size of the kernel such that the receptive field stays the same, i.e., is independent of the resolution $h$. In this section, we begin at the general graph neural operator and demonstrate how (considering the desirable properties of translation equivariance and efficiency) we arrive at local convolutional layers with adaptive kernel sizes using a fixed number of parameters. 

\paragraph{Graph neural operator (GNO):}
One of the most general instantiations of a neural operator is arguably the \emph{graph neural operator} (GNO)

\begin{align}
    \mathrm{GNO}_k[v](y) &=  \int_{U(y)} k(x, y) \cdot v(x) \, \mathrm{d}x \label{eq:gno} \\
    &\approx \sum_{x\in D^h \cap U(y)} k(x, y) \cdot v(x) \, q_x, \label{eq:gno_discr}
\end{align}
where $k$ is a kernel (typically parametrized by a neural network) and $q_x \in \mathbb{R}$ are suitable quadrature weights. While the GNO can represent local integral operators by picking a suitably small neighborhood $U(y) \subset D$, the evaluation of the kernel and aggregation in each neighborhood $U(y)$ is slow and memory-intensive for general kernels $k\colon D \times D \to \mathbb{R}^{n}$.
Moreover, for an arbitrary kernel, GNO is not equivariant w.r.t.\@ the symmetry group of the underlying domain. In particular, we lose the translation equivariance of convolutional layers for planar domains.

\paragraph{Fourier Neural Operator (FNO):} To retain translation equivariance, we can consider kernels of the form 
\begin{equation*}
    k(x,y) = \kappa(x - y)
\end{equation*}
and $U(y)=y+\operatorname{supp}(\kappa)$.
Then, we can rewrite the GNO as a convolution, i.e.,
\begin{equation}
\label{eq:gno_conv}
    \mathrm{GNO}_\kappa[v] = \kappa \star v.
\end{equation}
If we are dealing with periodic functions on the torus $D$, we can leverage the convolution theorem to compute~\eqref{eq:gno_conv}, i.e., 
\begin{equation}
    \mathcal{F}[\kappa \star v] = \overline{\mathcal{F}[\kappa]} \cdot \mathcal{F} [v],
\end{equation}
where $\mathcal{F}$ maps functions to their Fourier series coefficients. The \emph{Fourier Neural Operator} (FNO) now directly parametrizes $\overline{\mathcal{F}[\kappa]}$ and approximates $\mathcal{F}$ using the \emph{fast Fourier transform} given that $D^h$ is an equidistant grid.
While this leads to an efficient version of~\eqref{eq:gno_discr}, it assumes that $\overline{\mathcal{F}[\kappa]}$ has only finitely many nonzero Fourier modes---or, equivalently, that the kernel $\kappa$ has full support, making~\eqref{eq:gno_conv} a \emph{global} convolution.

\paragraph{Localized convolutions:}
To construct a locally supported kernel $\kappa$, we can directly discretize the convolution in~\eqref{eq:gno_conv}, i.e.,
\begin{equation}
\label{eq:gno_disco}
    \mathrm{GNO}_\kappa[v](y) \approx \sum_{x \in D^h} \kappa(x-y) \cdot v(x) \, q_x.
\end{equation}
Note that the sum can be taken only over the $x \in D^h$ with
\begin{equation*}
    x - y \in \operatorname{supp}(\kappa).
\end{equation*}
We remark that for an equidistant grid and constant\footnote{This is, e.g., the case for the trapezoidal rule on a torus.} quadrature weights, $q = q_x $, evaluating $\mathrm{GNO}_\kappa[v]$ at $y\in D^h$ corresponds to a standard convolution as in~\eqref{eq:conv} with kernel
\begin{equation*}
    K_{i} = q \kappa( z_i), \quad i=1,\dots,S,
\end{equation*}
where $z_i$ is defined as in \eqref{eq:conv_eval} and $S$ is sufficiently large such that $\operatorname{supp}(\kappa) \subset [z_1, z_S]$. See~\Cref{app:disco_1d} for details.
However, the advantage of the formulation in~\eqref{eq:gno_disco} is the fact that we can reuse the same kernel $\kappa$ across different resolutions (with the same receptive field $\operatorname{supp}(\kappa)$). 

The remaining question is centered around the parametrization of the kernel. One could draw inspiration from works on \emph{hypernetworks} in computer vision and parameterize it using a neural network~\cite{shocher2020discrete}. However, this is significantly more costly than a standard convolution. Another idea is to interpolate a fixed-sized kernel, e.g., using sinc or bi-linear interpolation. A more general version of the latter, based on a learnable linear combination of hat functions, is also known as \emph{discrete-continuous} (DISCO) convolution and has shown to be effective in different applications~\cite{Ocampo2022}. We note that this formulation is closest to the original convolutional layer typically used in computer vision. However, in the next section, we see that it also allows us to use unstructured meshes and formulate the operation on more general domains $D\subset \mathbb{R}^d$.

\subsection{General discrete-continuous convolutions}
\label{sec:disco}

The local support of the kernel in equation \eqref{eq:gno_disco} allows us to efficiently evaluate local convolutions in subdomains of $\mathbb{R}^d$ using sparse matrix-vector products. However, the operation $x-y$, which shifts the convolution kernel, is not well-defined on manifolds such as the sphere. To generalize the previous discussion to a Lie group $G$, we first replace the shift operator with the group action $g$ and obtain the so-called group convolution
\begin{align}
    \label{eq:group_conv}
    \operatorname{GroupConv}_\kappa[v](g) &= (\kappa \star v)(g) \nonumber\\
    &= \int_G \kappa(g^{-1}x) \cdot v(x)\;\mathrm{d}\mu(x),
\end{align}
where $g, x \in G$ and $\mathrm{d}\mu(x)$ is the invariant Haar measure on $G$. This formulation presents us with the challenge of being non-trivial to discretize. The framework of DISCO convolutions \cite{Ocampo2022} achieves this by approximating the integral with a quadrature rule while evaluating the group action continuously:
\begin{align}\label{eq:disco_conv}
    (k \star v)(g_i) &\approx \sum_{j=1}^m \kappa(g_i^{-1} x_j) \cdot v(x_j) \, q_j \nonumber\\
    &= \sum_{j=1}^m K_{ij} \cdot v(x_j) \, q_j.
\end{align}
Here $(x_j)_{j=1}^m \subset G$ represent quadrature points with corresponding quadrature weights $q_j$. For a given set of output positions $g_i$, we obtain the fully discrete formulation as matrix-vector product with the matrix $K_{ij} = \kappa(g_i^{-1} x_j)$, which is sparse due to the local support of the kernel. To parameterize the kernel, we choose a linear combination of basis functions $\kappa^{(\ell)}$, such that $\kappa = \sum_{\ell=1}^L \theta^{(\ell)}\kappa^{(\ell)}$ with trainable parameters $\theta^{(\ell)}\in\mathbb{R}$. 

Although we have presented the general idea only for Lie groups $G$, it is possible to construct the DISCO convolution also on manifolds with a group action acting on them, such as the 2-sphere $\mathbb{S}^2$. For a detailed discussion of DISCO convolutions and a construction in one dimension, we point the reader to~\Cref{app:disco} and \citet{Ocampo2022}.

\begin{remark}[Exact integration and equivariance]
    We note that DISCO convolutions satisfy equivariance properties for function classes that can be exactly integrated. For instance, on planar domains, these could be polynomials when using Legendre points. For equidistant grids on the torus, exact integration holds for bandlimited functions that are sampled above the Nyquist frequency. 
\end{remark}

\subsection{Local neural operator architecture}
\label{sec:architecture}
To design our neural operator, we want to combine pointwise, local, and global operations. To this end, we take an FNO (or SFNO for spherical problems) as a starting point, which already features global operations in Fourier space and pointwise operations using its residual connections. Then, we augment the operators by incorporating our proposed local convolutions from the previous section as additional branches in the respective layers (see Figure~\ref{fig:layer}). These four branches are summed pointwise within each local neural operator layer. The resulting architecture can be trained end-to-end with any standard operator learning training procedure and loss.

\section{Experiments}
\label{sec:experiments}
\begin{figure}[tb]
    \begin{subfigure}{.5\linewidth}
        \centering
        \begin{subfigure}{\linewidth}
            \centering
            \includegraphics[width=0.95\linewidth, trim={20px 20px 20px 20px}, clip]{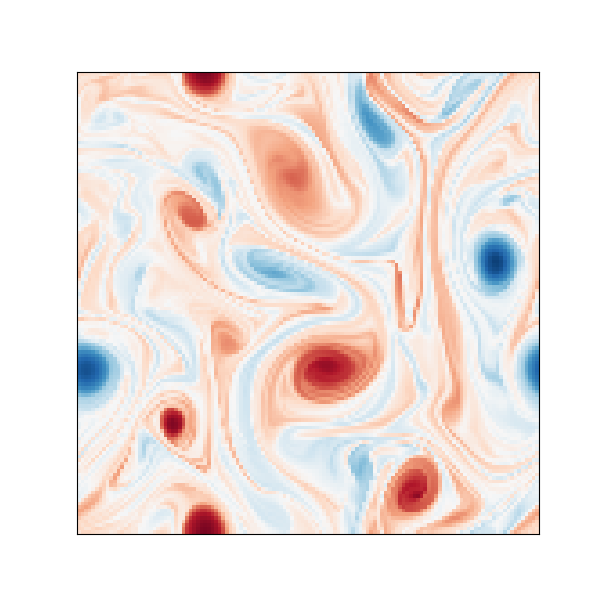}
            \caption*{initial condition}
        \end{subfigure}
        \begin{subfigure}{\linewidth}
            \centering
            \includegraphics[width=0.95\linewidth, trim={20px 20px 20px 20px}, clip]{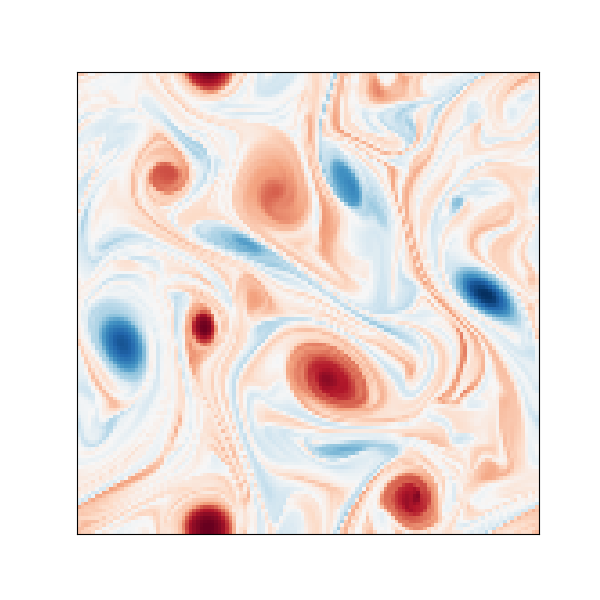}
            \caption*{ground truth}
        \end{subfigure} 
        \begin{subfigure}{\linewidth}
            \centering
            \includegraphics[width=0.95\linewidth, trim={20px 20px 20px 20px}, clip]{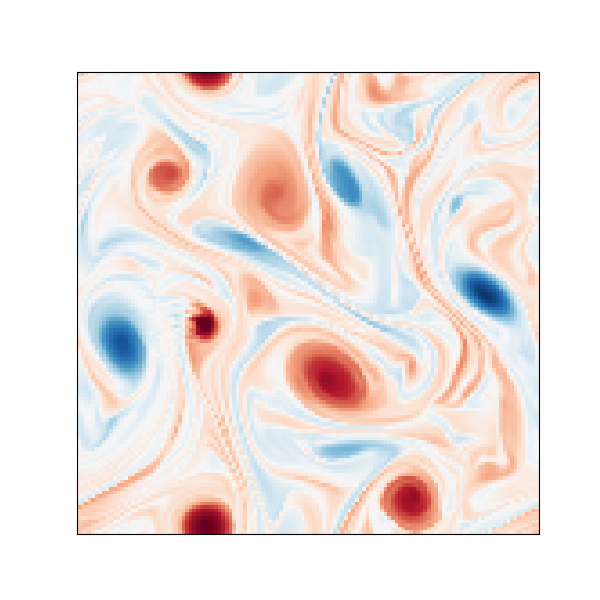}
            \caption*{prediction (5 steps)\vspace*{0.35em}}
        \end{subfigure}
        \caption{\textbf{Navier-Stokes Eqns.}}
    \end{subfigure}%
    \begin{subfigure}{.5\linewidth}
        \begin{subfigure}{\linewidth}
            \centering
            \includegraphics[width=0.95\linewidth, trim={20px 20px 20px 20px}, clip]{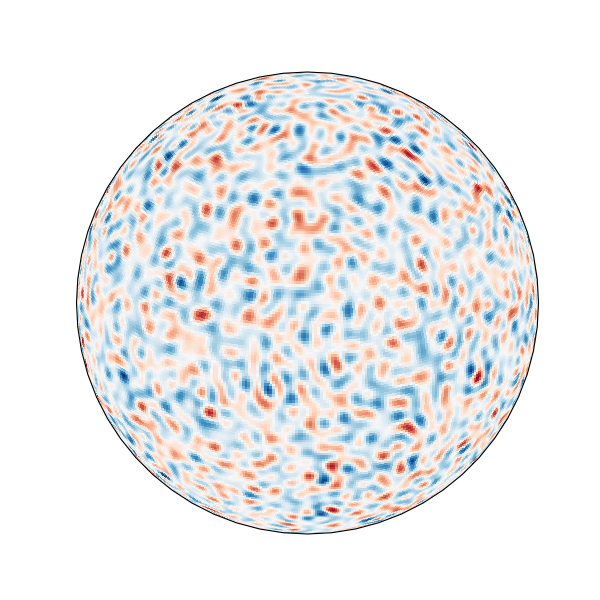}
            \caption*{initial condition}
        \end{subfigure}
        \begin{subfigure}{\linewidth}
            \centering
            \includegraphics[width=0.95\linewidth, trim={20px 20px 20px 20px}, clip]{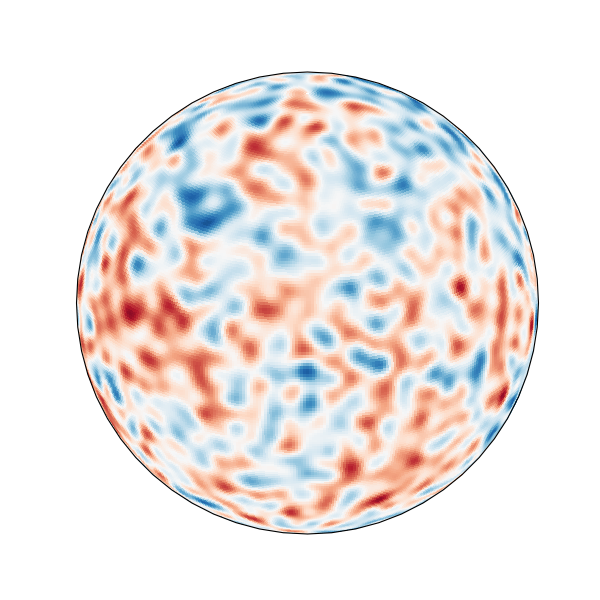}
            \caption*{ground truth}
        \end{subfigure}
        \begin{subfigure}{\linewidth}
            \centering
            \includegraphics[width=0.95\linewidth, trim={20px 20px 20px 20px}, clip]{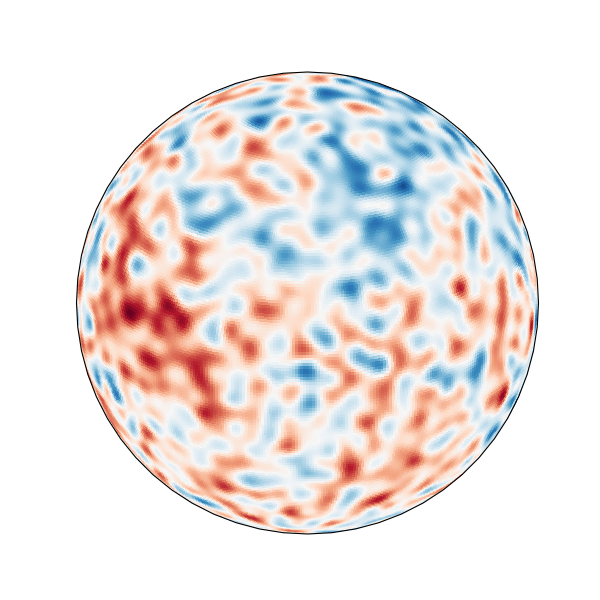}
            \caption*{prediction (5 steps)\vspace*{0.35em}}
        \end{subfigure}
        \caption{\textbf{Shallow Water Eqns.}}
    \end{subfigure}
    \caption{Initial condition, ground truth, and corresponding autoregressive predictions of our proposed models for the Navier-Stokes problem and the shallow water equations.}
    \label{fig:pde_examples}
    \vspace{-0.5em}
\end{figure}

\begin{figure*}[t!]
    \begin{subfigure}{0.5\linewidth}
        \centering
        \includegraphics[width=0.97\linewidth, trim={0px 0px 0px 0px}, clip]{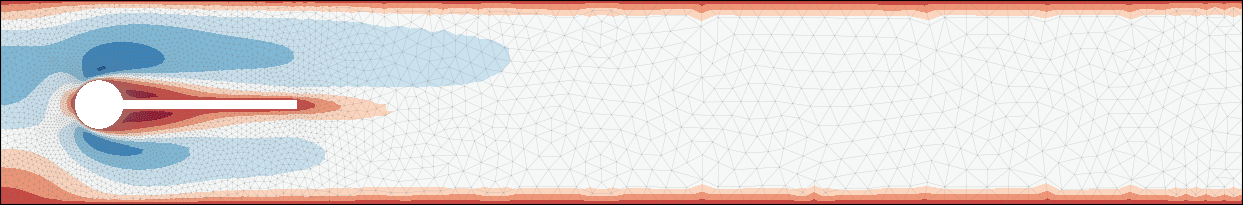}
        \caption*{ground truth}
    \end{subfigure}%
    \begin{subfigure}{0.5\linewidth}
        \centering
        \includegraphics[width=0.97\linewidth, trim={0px 0px 0px 0px}, clip]{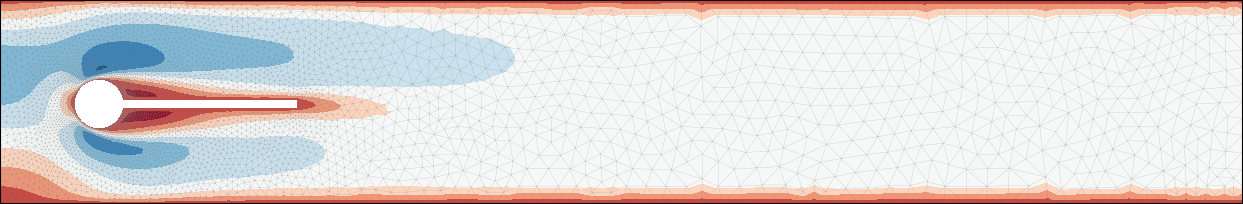}
        \caption*{prediction}
    \end{subfigure}
    \caption{Ground truth and prediction of the horizontal velocity for the flow past a cylinder. The data is represented on an unstructured mesh, which is visualized in gray color. Our proposed architecture can be readily applied to data on unstructured meshes such as this one. For this figure, we learn the residual to the previous time step.}
    \label{fig:cylinder_example}
    \vspace{-0.5em}
\end{figure*}

To validate the effectiveness of local operators, we evaluate the architecture on five PDE problems. \Cref{fig:pde_examples} and \Cref{fig:cylinder_example} show samples on three of these problems. In all five cases, we incorporate our proposed local operators into existing FNO and SFNO architectures to demonstrate that a significant improvement in performance can be achieved by introducing the inductive bias of local convolutions. This section describes the experimental setting for our problems. We provide further experiments, specific implementation details\footnote{Code is available in the \texttt{neuraloperator} library at \href{https://github.com/neuraloperator/neuraloperator}{github.com/neuraloperator/neuraloperator} and the \texttt{torch-harmonics} library at \href{https://github.com/NVIDIA/torch-harmonics}{github.com/NVIDIA/torch-harmonics}.}, and hyperparameters in~\Cref{sec:implementation_details}.

\subsection{Darcy flow}
\label{sec:darcy}
We first consider the steady-state, two-dimensional Darcy flow equation\begin{align}
\label{eq:darcy}
-\nabla\cdot (a  \nabla u)=f, \quad u|_{\partial D}=0,
\end{align}
on the domain $D=(0,1)^2$. In this problem, we motivate the need to incorporate local operators as inductive biases in neural operator architectures. To this end, we explicitly construct a problem that requires approximating the forcing function $f$ for a given diffusion coefficient $a$ and pressure $u$ in~\eqref{eq:darcy}. In other words, the task consists of learning the differential operator 
\begin{equation}
    \label{eq:darcy_ground_truth}
    u \mapsto -\nabla\cdot (a  \nabla u).
\end{equation} 
Following the setup of~\citet{hasani2024generating}, we take 
\begin{equation*}
    a(x) = 
    \begin{bmatrix}
        x_1^2 & \sin(x_1x_2) \\
        x_1 + x_2 & x_2
    \end{bmatrix}
\end{equation*}
for $x \in D$. We then compare the performance of our proposed models: FNO in parallel with differential kernels, FNO with integral kernels, and FNO with both kernels. We also compare with a baseline FNO and the U-Net architecture from~\citet{gupta2022towards}. Moreover, we perform zero-shot super-resolution on this problem; these results can be found in~\Cref{sec:superres}.

\subsection{Navier-Stokes equations}
\label{sec:ns}
Next, we consider the two-dimensional Navier-Stokes equation on the torus. In particular, we consider Kolmogorov flows, which can be described by
\begin{align}
\label{eq:navier_stokes}
\partial_t u + u \cdot \nabla u - \frac{1}{\textrm{Re}} \Delta u &=  - \nabla p + \sin(my) \hat x \\
\nabla \cdot u &= 0 
 \nonumber \end{align}
on the spatial domain $D=(0,2\pi)^2$ equipped with periodic boundary conditions. In the above, $u$ and $p$ denote the velocity and pressure, and $\textrm{Re}$ denotes the Reynolds number of the flow. We want to learn the solution operator mapping initial conditions $u(\cdot, 0) = u_0$ to the time-evolved solution in vorticity form $w(\cdot, \tau)$ at time $\tau\in(0,\infty)$, where the vorticity is given by \begin{equation}
    w = (\nabla \times u) \hat z
\end{equation}
with $\hat z$ being a unit vector normal to the plane.

We consider $m=4$ and $\text{Re} = 5000$, and we emphasize that learning the solution operator for such a high Reynolds number is very challenging due to the turbulent nature and small-scale features of the flow. We conjecture that the baseline FNO is prone to over-smoothing over the finer scales, consequently leading to a degradation of performance. 

To validate this conjecture, we compare the performance of the local neural operator with a baseline FNO and U-Net and evaluate the effectiveness of our proposed differential and integral kernels, respectively.

\subsection{Diffusion-Reaction equation}

Moreover, we consider the 2D Diffusion-Reaction equation from~\citet{takamoto2022pdebench} which can, for instance, be used for modeling biological pattern formation. In particular, we want to predict the time-evolution of two non-linearly coupled variables, i.e., the activator $u$ and the inhibitor $v$, solving the equation
\begin{align}
       \partial_t u &= c_u \Delta u + R_u(u,v), \\
       \partial_t v &= c_v \Delta v + R_v(u,v),
\end{align}
for the spatial domain $D = (-1,1)^2$ with no-flow Neumann boundary condition. The reaction functions $R_u$ and $R_v$ are given by the Fitzhugh-Nagumo equation
\begin{equation}
    R_u(u,v) = u - u^3 -k - v, \quad R_v(u,v) = u-v
\end{equation}
and we consider $c_u = 10^{-3}$, $c_v = 5 \cdot 10^{-3}$, and $k = 5 \cdot 10^{-3}$. The task is to learn the operator mapping the state $(u,v)$ at consecutive time-steps to the time-evolved state and we can compare against the baselines from~\citet{takamoto2022pdebench}.

\begin{table*}[t!]
\vskip 3pt
\caption{Results for the 2D diffusion-reaction equation from PDEBench~\cite{takamoto2022pdebench}. We present their U-Net baseline results and metrics. In particular, we report the RMSE errors for the low, medium, and high frequency ranges (fRMSE low/med./high), and the (absolute) RMSE, the maximum error, RMSE errors at the boundary (bRMSE), as well as RMSE for the conserved values (cRMSE).}
\label{tab:diff_reac}
\vspace{-0.75em}
\begin{center}
\setlength\tabcolsep{3.5pt}
\resizebox{\linewidth}{!}{\begin{tabular}{lccccccccc}
    \toprule
    Model & 
    \@\# Params & 	Rel. $L^2$-error & fRMSE low & fRMSE med.\@ & fRMSE high & RMSE & max. error & bRMSE & cRMSE\\
    \midrule
    U-Net & 
    $7.8 \cdot 10^5$  &  $8.4 \cdot 10^{-1}$ & $1.7 \cdot 10^{-2}$ & $8.2 \cdot 10^{-4}$ & 	$5.7 \cdot 10^{-2}$ & $6.1 \cdot 10^{-2}$ & $1.9\cdot 10^{-1}$ & $7.8 \cdot 10^{-2}$ & $3.9 \cdot 10^{-2}$ \\
    FNO & 
    $9.3 \cdot 10^5$ & $8.3 \cdot 10^{-2}$ & $6.2 \cdot 10^{-4}$ & $5.6 \cdot 10^{-4}$ & $2.4 \cdot 10^{-4}$ & $5.2 \cdot 10^{-3}$ & $7.3 \cdot 10^{-2}$ & $1.5 \cdot 10^{-2}$ & $1.2 \cdot 10^{-3}$ \\
    \textbf{FNO + loc. int (ours)} & 
    $8.8 \cdot 10^5$ & $6.3 \cdot 10^{-2}$ & $4.0 \cdot 10^{-4}$ & $4.6 \cdot 10^{-4}$ & $1.5 \cdot 10^{-4}$ & $3.6 \cdot 10^{-3}$ & $5.0 \cdot 10^{-2}$ & $1.0 \cdot 10^{-2}$ & $\mathbf{4.8 \cdot 10^{-4}}$ \\
    \textbf{FNO + diff. (ours)} & 
    $8.8 \cdot 10^5$ & $3.4 \cdot 10^{-2}$ & $4.4 \cdot 10^{-4}$ & $1.9\cdot 10^{-4}$ & $\mathbf{6.1\cdot 10^{-5}}$ & $1.9 \cdot 10^{-3}$ & $3.5 \cdot 10^{-2}$ & $4.4 \cdot 10^{-3}$ & $1.1\cdot 10^{-3}$ \\
    \textbf{FNO + loc. int. + diff. (ours)} & 
    $8.9 \cdot 10^5$ & $\mathbf{3.1 \cdot 10^{-2}}$ & $\mathbf{3.2 \cdot 10^{-4}}$ & $\mathbf{1.8\cdot 10^{-4}}$ & $6.2\cdot 10^{-5}$ & $\mathbf{1.7 \cdot 10^{-3}}$ & $\mathbf{3.3 \cdot 10^{-2}}$ & $\mathbf{4.2 \cdot 10^{-3}}$ & $7.4\cdot 10^{-4}$ \\
    \bottomrule
\end{tabular}}
\end{center}
\end{table*}

\subsection{Shallow water equations}
\label{sec:swe}
To test the approach on the sphere, we consider the shallow water equations on the rotating sphere. They represent a system of hyperbolic partial differential equations used to model a variety of geophysical flow phenomena, such as atmospheric flows, tsunamis, and storm surges. They can be formulated in terms of the evolution of two state variables $\varphi$ and $u$ (geopotential height and the tangential velocity of the fluid column), governed by the equations
\begin{gather}
\begin{aligned}
\partial_t\varphi + \nabla \cdot (\varphi u) = 0, \\ 
\partial_t(\varphi u) + \nabla \cdot F = f, 
\end{aligned}
\end{gather}
on the sphere $D= \mathbb{S}^2$ with suitable initial conditions $\varphi(\cdot, 0)=\varphi_0$ and $u(\cdot, 0) = u_0$,
a momentum flux tensor
\begin{equation}
    F_{ij} = \varphi u_i u_j + \frac{1}{2} \varphi^2,
\end{equation}
and a source term 
\begin{equation}
    f = - 2 \Omega x \times (\varphi u),
\end{equation}
which models the Coriolis force due to the rotation of the sphere with angular velocity $\Omega$.

As baselines, we use a planar U-Net architecture, a spherical U-Net, where convolutions are performed using local integral kernels on the sphere, and an SFNO architecture. On the sphere, we only consider the impact of the integral kernel on the SFNO architecture, as it allows for a natural extension to the sphere (see~\Cref{sec:disco}).

\subsection{Flow past a cylinder}
\label{sec:cylinder_flow}
Due to the formulation of the discrete-continuous convolutions, our approach can be readily generalized to unstructured meshes. To demonstrate this, we consider the Navier-Stokes equations in a two-dimensional channel with a cylindrical structure and a membrane attached to it \citep[see][]{rahman2024pretraining}. \Cref{fig:cylinder_example} depicts a sample from the validation dataset alongside predictions from our model.

\begin{table}[tb]
    \vskip 3pt
    \setlength\tabcolsep{1.9pt}
    \caption{Results for the cylinder flow problem in a low-data regime with 250 training samples. Baseline results are taken from \citet{rahman2024pretraining} and all architecture including ours perform a direct prediction (no residual prediction) of the velocity and pressure at the next time step.}
    \label{tab:cylinder}
    \vspace{-0.75em}
    \begin{center}
    \addtolength{\tabcolsep}{2pt}
    \resizebox{\linewidth}{!}{\begin{tabular}{lcc}
    \toprule
    Model & \# Parameters & MSE error\\
    \midrule
    GINO      & $6 \cdot 10^7$  & $2.09 \cdot 10^{-2}$  \\
    DeepONet  & $6 \cdot 10^6$   & $1.39 \cdot 10^{-1}$ \\
    GNN       & $6 \cdot 10^5$ & $5.00 \cdot 10^{-3}$  \\
    ViT       & $3 \cdot 10^7$  & $1.19 \cdot 10^{-2}$  \\
    U-net     & $3 \cdot 10^7$  & $9.34 \cdot 10^{-2}$  \\
    \textbf{FNO + local integral kernel (ours)}  & $10 \cdot 10^6$  & $\mathbf{2.88 \cdot 10^{-3}}$  \\
    \bottomrule
\end{tabular}}
\end{center}
\vspace{-1em}
\end{table}

The dataset considers the flow past the cylinder with Reynolds number $\text{Re}=2000$. \citet{rahman2024pretraining} provide several baselines for a low-data regime (250 samples), where the task is to predict the velocity and pressure after a given time step.
To deal with the unstructured grid, we adapt the architecture to utilize local integral operators as encoders and decoders to transform the data into a latent representation on an equidistant grid. Then, we can use an FNO with our local integration layers in the latent space.

\begin{table*}[tb!]
    \vskip 3pt
    \caption{Experimental results for Darcy flow, Navier-Stokes, and the spherical shallow water problems. For all three problems, the test error is reported in terms of the relative $L^2$-error after a single step. For the time-dependent Navier-Stokes and shallow water equations, we also predict the error after $5$ autoregressive steps.}
    \vspace{-0.75em}
    \label{tab:results}
    \begin{center}
    \begin{small}
    \addtolength{\tabcolsep}{2pt}
    \begin{tabular}{lcccccc}
    \toprule
    \multirow{2}{*}[-1pt]{Model} & \multicolumn{4}{c}{Parameters} & \multicolumn{2}{c}{Relative $L^2$-Error} \\
    \cmidrule(lr){2-5} \cmidrule(lr){6-7}
     & \scriptsize{\# Layers} & \scriptsize{\# Modes} & \scriptsize{Embedding} & \scriptsize{\# Parameters}  & \scriptsize{1 step} & \scriptsize{5 steps} \\
    \midrule
    \multicolumn{7}{c}{Darcy Flow} \\
    \midrule
    U-Net            & 17 & - & 18 & $2.850 \cdot 10^6$ & $1.380 \cdot 10^{-2}$ & - \\
    FNO              & 4 & 20 & 41 & $2.715 \cdot 10^6$ & $5.867 \cdot 10^{-2}$ & -  \\
    \textbf{FNO + diff. kernel (ours)} & 4 & 12 & 65 & $2.638 \cdot 10^6$ & $\mathbf{7.357 \cdot 10^{-3}}$ & -  \\
    FNO + local integral kernel (ours) & 4 & 20 & 40 & $2.617 \cdot 10^6$ & $6.034 \cdot 10^{-2}$ & - \\
    FNO + local integral + diff. kernel (ours) & 4 & 12 & 64 & $2.639 \cdot 10^6$ & $9.032 \cdot 10^{-3}$ & - \\
    \midrule
    \multicolumn{7}{c}{Navier-Stokes Equations} \\
    \midrule
    U-Net            & 17 & - & 56 & $2.758 \cdot 10^7$ & $1.674 \cdot 10^{-1}$ & $5.115 \cdot 10^{-1}$ \\
    FNO              & 4 & 40 & 65 & $2.711 \cdot 10^7$ & $1.381 \cdot 10^{-1}$ & $2.360 \cdot 10^{-1}$ \\
    FNO + diff. kernel (ours) & 4 & 40 & 65 & $2.726\cdot 10^7$ & $1.073 \cdot 10^{-1}$ & $2.129 \cdot 10^{-1}$ \\
    FNO + local integral kernel (ours) & 4 & 20 & 129 & $2.716 \cdot 10^7$ & $1.110 \cdot 10^{-1}$ & $2.183 \cdot 10^{-1}$ \\
    \textbf{FNO + local integral + diff. kernel (ours)} & 4 & 20 & 127 & $2.691 \cdot 10^7$ & $\mathbf{9.022 \cdot 10^{-2}}$ & $\mathbf{1.956 \cdot 10^{-1}}$ \\
    \midrule
    \multicolumn{7}{c}{Spherical Shallow Water Equations} \\
    \midrule
    U-Net & 17 & - & 32 & $2.898 \cdot 10^6$ & $1.341 \cdot 10^{-3}$ & $1.226 \cdot 10^{-2}$\\
    Spherical U-Net (with local integral kernel) & 17 & - & 32 & $1.639 \cdot 10^6$ & $6.160 \cdot 10^{-4}$ & $3.265 \cdot 10^{-3}$\\
    SFNO             & 4 & 128 & 32 & $1.066 \cdot 10^6$ & $9.220 \cdot 10^{-4}$ & $3.185 \cdot 10^{-3}$ \\
    \textbf{SFNO + local integral kernel (ours)} & 4 & 128 & 31 & $1.019 \cdot 10^6$ & $\mathbf{2.624 \cdot 10^{-4}}$ & $\mathbf{5.392 \cdot 10^{-4}}$ \\
    \bottomrule
    \end{tabular}
    \addtolength{\tabcolsep}{-2pt}
    \end{small}
    \end{center}
\vspace{-0.5em}
\end{table*}

\subsection{Results and discussion}
The results of our numerical experiments are reported\footnote{In all settings, hyperparameters are chosen to result in macro-architectures with similar parameter counts for the purpose of comparability. The experimental setup, including the choice of hyperparameters, is outlined in \Cref{sec:implementation_details} in the Appendix.} in~\Cref{tab:results,tab:cylinder,tab:diff_reac}. We observe significant performance gains over the baselines in all five problem settings. In particular, we notice that the best performance in the reported relative $L^2$-errors is achieved with the inclusion of both global and local operations, despite an overall reduction in parameter count with respect to their corresponding FNO/SFNO baselines. This is particularly pronounced for the Navier-Stokes and shallow water equations, where errors can accumulate in autoregressive roll-outs. We attribute this to our models' inductive bias, allowing it to better capture the fine-grained scales and thus achieve better performance. In~\Cref{tab:diff_reac}, we also show that our local layers achieve lower errors at high frequencies for the diffusion-reaction
equation, indicating that local, high-frequency features are correctly captured.

The Navier-Stokes problem demonstrates the efficacy of combining both differential and local integral kernels with the global convolution of the FNO. Our proposed model with these three components together outperforms models with only two of these three operations. While hyperbolic PDEs likely require only local receptive fields (due to finite information propagation~\citep{LeVeque1992}) and elliptic problems instead require global information, we observe that many practical problems involving a mixture of operators benefit from a hybrid approach such as ours.

We note that the Darcy flow problem is meant to motivate the need for our proposed differential kernels (and indeed, our best-performing model achieves $87\%$ lower relative $L^2$-error than FNO). Since the ground truth operator in \eqref{eq:darcy_ground_truth} is a differential operator, we expect (and observe) that for high accuracies, very local (i.e., differential) kernels are needed. In particular, we see that the baseline FNO performs poorly, and the local integral kernels do not provide additional benefits. However, we include them in~\Cref{tab:results} for completeness.

Finally, we perform super-resolution experiments, in which we train the model at a given resolution and then evaluate it at another resolution without fine-tuning. \Cref{fig:superres} depicts super-resolution results at twice the training resolution for both the Darcy and shallow water equations. The corresponding numerical results are listed in \Cref{tab:superres}. We observe that our approach generalizes well across resolutions and outperforms the baselines across all tested resolutions. For a detailed discussion of the super-resolution experiments, we refer the reader to \Cref{sec:superres}.
\section{Conclusion}
\label{sec:conclusion}

In this paper, we have demonstrated a novel framework for local neural operators. We have shown how convolutional layers can be constrained to realize neural operators that approximate differential operators in the continuous limit.
Moreover, we have derived convolutions with local integral kernels from the general notion of an integral transform and the related graph neural operator.
Finally, we have constructed localized neural operators on the sphere by using discrete-continuous convolutions \cite{Ocampo2022}.

The resulting neural operators introduce a strong inductive bias for learning operators with local receptive fields. In particular, their formulation ensures the same local operation everywhere in the domain. This equivariance (w.r.t.\@ the underlying symmetry group) reduces the required number of learnable parameters and improves generalization.

Our numerical experiments demonstrate consistent improvements when existing neural operators with global receptive fields are augmented with the proposed localized convolutions, resulting in reductions in relative $L^2$-error of up to 72\% over the corresponding baselines. We thus expect local neural operators to play an important role in solving real-world scientific computing problems with machine learning.

\section*{Acknowledgements}
The authors thank Md Ashiqur Rahman for useful discussions about benchmarks and baselines. M. Liu-Schiaffini is grateful for support from the Mellon Mays Undergraduate Fellowship. J. Berner acknowledges support from the Wally Baer and Jeri Weiss Postdoctoral Fellowship. A. Anandkumar is supported in part by Bren endowed chair and by the AI2050 senior fellow program at Schmidt Sciences. 
\section*{Impact Statement}
The aim of this work is to advance the field of machine learning and scientific computing. While there are many potential societal consequences, none of them are immediate to require specifically being highlight here.
\bibliography{references}
\bibliographystyle{icml2024}

%%%%%%%%%%%%%%%%%%%%%%%%%%%%%%%%%%%%%%%%%%%%%%%%%%%%%%%%%%%%%%%%%%%%%%%%%%%%%%%
%%%%%%%%%%%%%%%%%%%%%%%%%%%%%%%%%%%%%%%%%%%%%%%%%%%%%%%%%%%%%%%%%%%%%%%%%%%%%%%
% APPENDIX
%%%%%%%%%%%%%%%%%%%%%%%%%%%%%%%%%%%%%%%%%%%%%%%%%%%%%%%%%%%%%%%%%%%%%%%%%%%%%%%
%%%%%%%%%%%%%%%%%%%%%%%%%%%%%%%%%%%%%%%%%%%%%%%%%%%%%%%%%%%%%%%%%%%%%%%%%%%%%%%
\newpage
\appendix
\onecolumn

\section{General differential kernels}

\subsection{Theoretical construction}
\label{app:diff}
In the following, we present the general idea for irregular grids, from which~\Cref{prop:diff} follows as a special case. Let us first define the assumptions on our grids.
\paragraph{Regular discrete refinement:}
Let $\|\cdot\|$ be a norm on $\mathbb{R}^d$ and denote by $B_h(x)\subset \mathbb{R}^d$ the ball of radius $h\in(0,\infty)$ around $x\in\mathbb{R}^d$ w.r.t.\@ $\|\cdot\|$. Further, let $D\subset \mathbb{R}^d$ be a domain and let $(h_\ell)_{\ell\in\mathbb{Z}}$ be a sequence that converges to zero. We call $(D_\ell)_{\ell\in\mathbb{Z}}\subset D$ a \emph{regular discrete refinement} with widths $(h_\ell)_{\ell\in\mathbb{Z}}$ if there exists $N\in\mathbb{N}$ such that for all $\ell \in \mathbb{Z}$ and $x\in\mathbb{R}^d$ we have that
\begin{equation*}
    |B_{h_\ell}(x) \cap D_\ell| \le N \quad \text{and} \quad  \operatorname{span}\left(\left\{ \begin{bmatrix} 1 \\ y  \end{bmatrix} - \begin{bmatrix} 1 \\ x  \end{bmatrix} \colon y \in B_{h_\ell}(x) \cap D_\ell \right\} \right) = \mathbb{R}^{d+1}.
\end{equation*}
The second assumption states that we can find an \emph{affinely independent} subset in each ball. Note that, for instance, equidistant grids satisfy these assumptions.

\paragraph{First-order differential operator:}
We want to find bounded kernels $k(x,y)\colon \mathbb{R}^d \times \mathbb{R}^d \to \mathbb{R}$ with the following property: 
There exist $c\in\mathbb{R}$ and $b\in \mathbb{R}^d$ such that for all $v\in C^1(D,\mathbb{R})$, all $y\in\mathbb{R}^d$, and any regular discrete refinement $(D_\ell)_{\ell\in\mathbb{Z}}\subset \mathbb{R}^d$ with widths $(h_\ell)_{\ell\in\mathbb{Z}}$ it holds that
\begin{equation}
\label{eq:def}
    \lim_{\ell \to \infty} \ \sum_{x \in B_{h_\ell}(y) \cap D_\ell} k(x,y) v(x) = c v(y) + \nabla v(y) \cdot b.
\end{equation}
We will now investigate which additional assumptions are needed.
Let us fix $y\in\mathbb{R}^d$ and enumerate 
\begin{equation*}
   (x_j)_{j=1}^m \coloneqq  B_{h_\ell}(y) \cap D_\ell.
\end{equation*}
Then, we can use Taylor's theorem to show that
\begin{align}
    \sum_{x \in B_{h_\ell}(y) \cap D_\ell} k(x,y) v(x) &= \sum_{j=1}^m k(x_j, y) \left( v(y) +  \nabla v(y) \cdot (x_j - y) + \mathcal{O}( h_\ell) \right) \\
    & = v(y) \sum_{j=1}^m  k(x_j, y)  + \nabla v(y) \cdot \left(\sum_{j=1}^m  k(x_j, y) (x_j - y)\right)   + \mathcal{O}( m \|k\|_{L^\infty} h_\ell).
\end{align}
Since the kernel and $m$ are bounded (uniformly over $\ell\in\mathbb{N}$), we have that $\mathcal{O}( m \|k\|_{L^\infty} h_\ell)=\mathcal{O}( h_\ell)$. As we want to guarantee convergence to $c v(y) + \nabla v(y) \cdot b$ for suitable $c\in \mathbb{R}$ and $b \in\mathbb{R}^d$ (independent of $y$), we need to satisfy that
\begin{equation}
    c = \sum_{j=1}^m k(x_j, y) 
\end{equation}
and that
\begin{equation}
    b = \sum_{j=1}^m  k(x_j, y) (x_j - y) 
\end{equation}
This yields the linear\footnote{For fixed $y$.} system 
\begin{equation}
\label{eq:lin_sys}
   \underbrace{\begin{bmatrix} 1 & \dots & 1 \\ x_1 - y  & \dots &   x_m - y \end{bmatrix}}_{\in \mathbb{R}^{(d + 1) \times n}} \begin{bmatrix}  k(x_1, y) \\ \vdots \\   k(x_m, y) \end{bmatrix} = \begin{bmatrix} c \\ b \end{bmatrix}.
\end{equation}
Our assumptions on the refinement guarantee that we can find linearly independent columns such that we can solve the system. However, generally and by abuse of notation, the value of $k(x_j,y)$ depends on all the points $(x_j)_{j=1}^m$. However, for an equidistant grid, one can directly see that the convolutional kernel given in~\Cref{prop:diff} satisfies~\eqref{eq:lin_sys} with $c=0$.

\subsection{Empirical evaluation}
\begin{figure}[t]
    \centering
    \begin{subfigure}{.19\linewidth}
        \centering
        \includegraphics[width=0.95\linewidth, trim={20px 37px 0px 20px}]{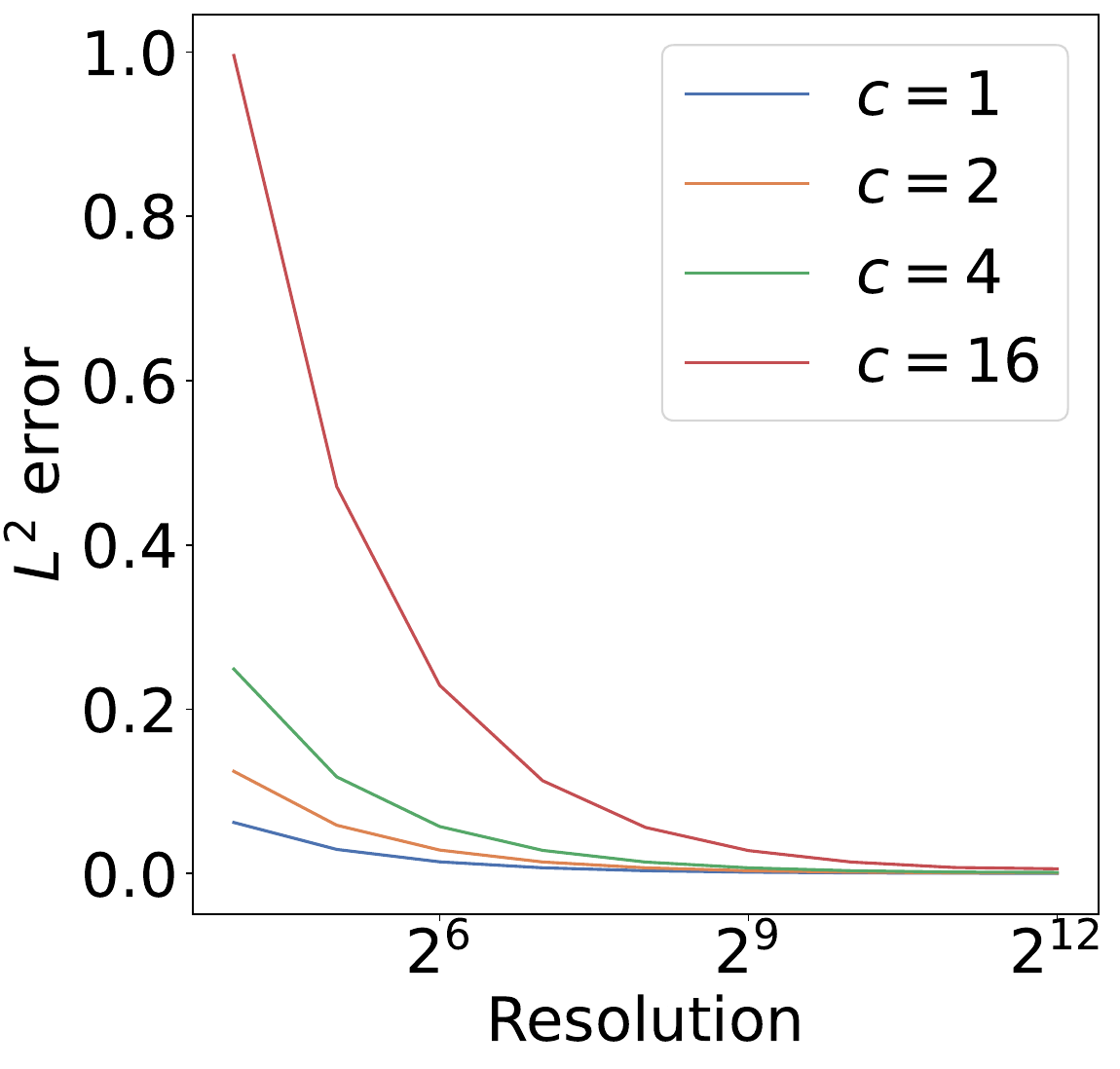}
        \caption{Convergence}
    \end{subfigure}
    \begin{subfigure}{.19\linewidth}
        \centering
        \includegraphics[width=0.95\linewidth, trim={20px 20px 20px 20px}, clip]{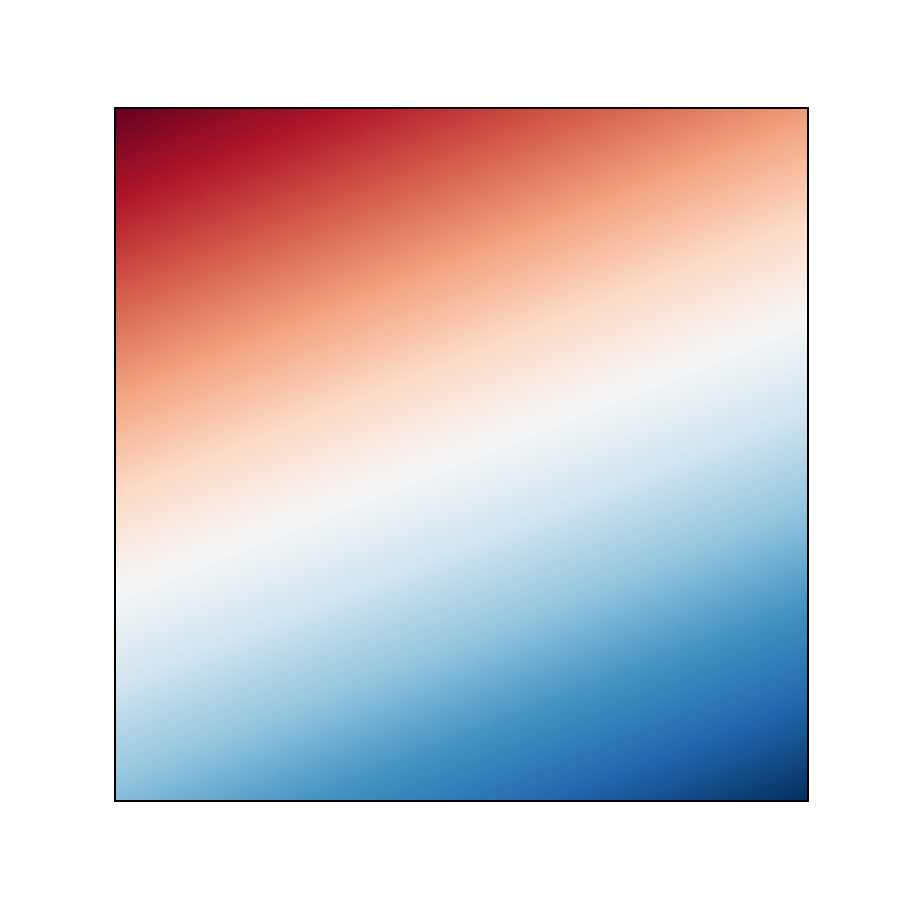}
        \caption{Ground truth}
    \end{subfigure}
    \begin{subfigure}{.19\linewidth}
        \centering
        \includegraphics[width=0.95\linewidth, trim={20px 20px 20px 20px}, clip]{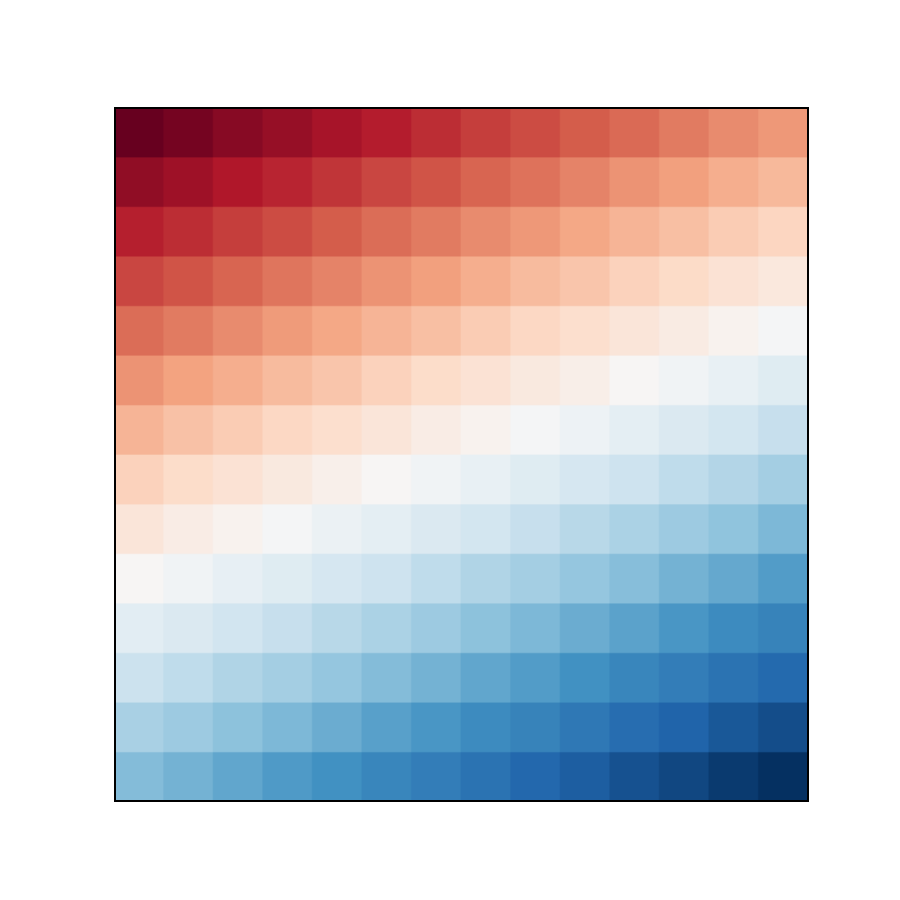}
        \caption{Low-res prediction}
    \end{subfigure}
    \begin{subfigure}{.19\linewidth}
        \centering
        \includegraphics[width=0.95\linewidth, trim={20px 20px 20px 20px}, clip]{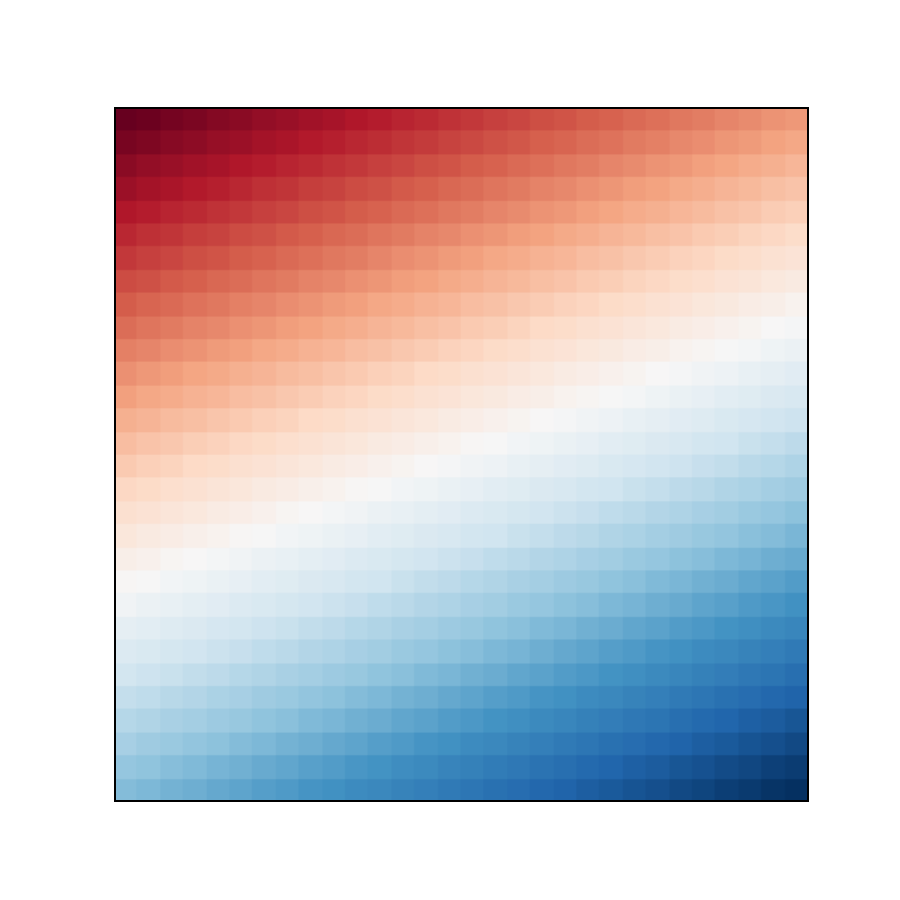}
        \caption{Medium-res prediction}
    \end{subfigure}
    \begin{subfigure}{.19\linewidth}
        \centering
        \includegraphics[width=0.95\linewidth, trim={20px 20px 20px 20px}, clip]{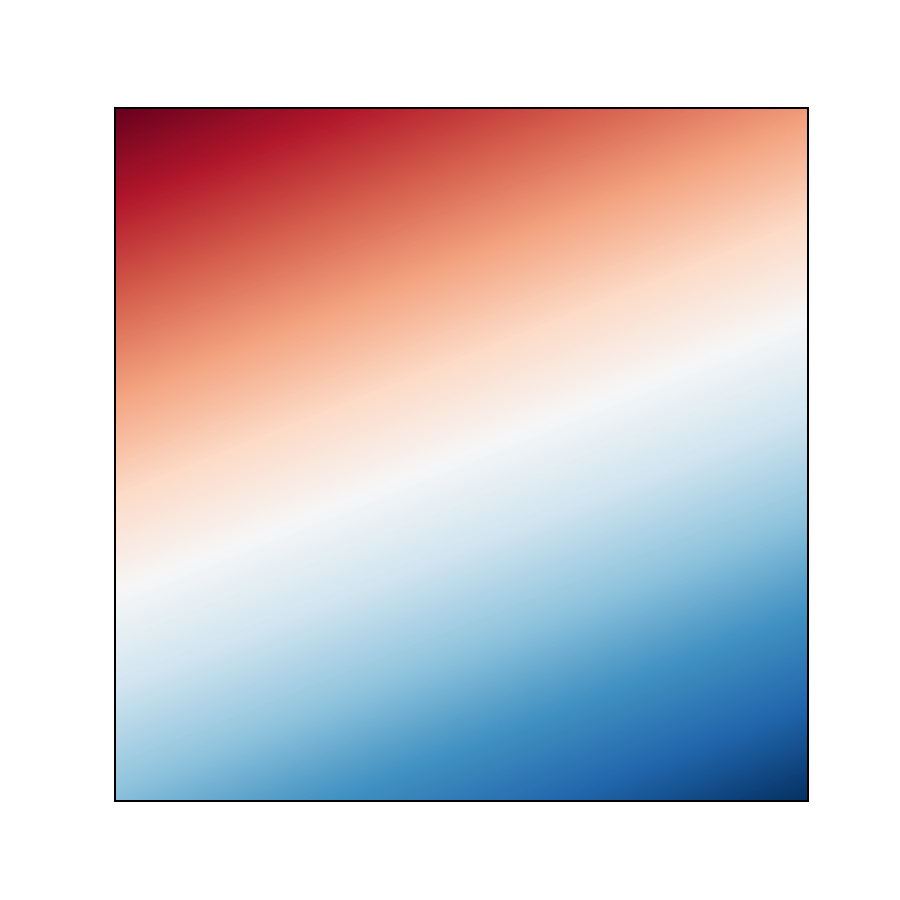}
        \caption{High-res prediction}
    \end{subfigure}
    \caption{Empirical evaluation of the proposed differential kernel. (a) $L^2$ errors at various resolutions and quadratic coefficient scales $c$. (b) True differential operator for $c = 1$. (c) Output of the differential kernel at a resolution of $32 \times 32$. (d) Output of the differential kernel at a resolution of $64 \times 64$. (e) Output of the differential kernel at a resolution of $4096 \times 4096$.}
    \label{fig:parabola_fig}
\end{figure}

\label{app:diff_empirical}
In this section, we empirically evaluate the convergence of our proposed differential kernels to a differential operator as the resolution increases. In particular, we consider a randomly-initialized differential kernel satisfying the constraints discussed in~\Cref{sec:diff_layer} (mean zero and scaled by the resolution $\frac{1}{h}$, where $h$ is the width of each grid cell in an equidistant grid).

To be able to compute the theoretical differential operator in closed form, we consider a simple parabola $v\colon [0,1]^2 \to \mathbb{R}^n$, given by
\begin{equation}
    \label{eq:parabola}
    x \mapsto \|x\|^2 \begin{bmatrix} c_1 ,\dots, c_n\end{bmatrix}^\top,
\end{equation}
with coefficients $c_1, \dots, c_n$.

We define a randomly-initialized first-order differential kernel $K=(K_{ij})_{i,j=1}^{3,3} \subset \mathbb{R}^{n}$ (i.e., one differential kernel for each of the $n$ input channels) subject to the constraints described in~\Cref{sec:diff_layer}. Let $b_j\in\mathbb{R}^n$ denote the direction corresponding to the directional derivative corresponding to the differential kernel for channel $j$, see~\Cref{app:diff}. With the parabola as defined in~\eqref{eq:parabola}, the output function of the true differential operator corresponding to $K$ is given by
\begin{equation}
    x \mapsto 2 \sum_{i=j}^n x \cdot b_j.
\end{equation}

In this simple experiment, we set $n = 10$ and generate $c_1, \dots, c_n$ uniformly on the unit interval, multiplied by some scaling factor $c \in \{1, 2, 4, 16\}$. We discretize the parabola in~\eqref{eq:parabola} over different resolutions, 
convolve the kernel $K$ with each of these discretizations, and compute the $L^2$ error with respect to the true differential operator. \Cref{fig:parabola_fig} shows the $L^2$ error for these values of $c$ and resolutions, as well as visualizations of the true differential operator and the one outputted by the differential kernel at three different resolutions.

From~\Cref{fig:parabola_fig}, we observe a clear convergence to the differential operator as the resolution increases. For a fixed resolution, convergence is slower for greater magnitudes of the second derivative (i.e., larger $c$). This can be theoretically shown by estimating the remainder of the Taylor expansion in~\Cref{app:diff}. 
\section{Discrete-continuous convolutions}
\label{app:disco}

\subsection{General Ideas}

The following section outlines the basic ideas behind discrete-continuous convolutions as introduced by \cite{Ocampo2022}. To generalize the (continuous) convolution \eqref{eq:gno_conv} to Lie groups and quotient spaces of Lie Groups, we consider the group convolution (see e.g. \citet{Cohen2016}).
\begin{definition}[Group Convolution]
Let $\kappa, v: G \rightarrow \mathbb{R}$ be functions defined on the group $G$. The group convolution is given by
\begin{equation}
    \tag{\ref{eq:group_conv}}
    (\kappa \star v)(g) = \int \kappa(g^{-1}x) \cdot v(x) \,\mathrm{d}\mu(x),
\end{equation}
where $g, x \in G$ and $\mathrm{d}\mu(x)$ is the invariant Haar measure on G.
\end{definition}

\begin{remark}
In some cases, signals are not defined on a group but rather on a quotient space $G/H$, where $H$ is a subgroup of $G$. In such cases, a convolution may still be defined by taking $g \in G/H$. For an example, see spherical convolutions \cite{Driscoll1994, Ocampo2022}.
\end{remark}

While group convolutions can typically be computed by generalized Fourier transforms on the corresponding manifolds, their usage is generally preferred if the convolutions are non-local operators, i.e., the convolution kernel $\kappa$ is not compactly supported. On the periodic domain $\mathbb{T}^d$ (i.e., Euclidean space with periodic boundaries), such convolutions are typically computed discretely by directly sliding the kernel. 

\begin{definition}[DISCO convolutions]
Given a quadrature rule with quadrature points $x_j \in G$ and quadrature weights $q_j$, we approximate the group convolution \eqref{eq:group_conv} with the discrete sum
\begin{equation}
    (\kappa \star v)(g) = \int \kappa(g^{-1} x) \cdot v(x)\,\mathrm{d}\mu(x) \approx \sum_{j=1}^m \kappa(g^{-1} x_j) \cdot v(x_j) \, q_j.
\end{equation}
In particular, the group action $g$ is applied analytically to the kernel function $\kappa$, whereas the integral is approximated using the quadrature rule.
\end{definition}

For a discrete set of output locations $g_i$, this becomes a straight-forward matrix-vector multiplication
\begin{align}
    \label{eq:disco_convolution}
    \sum_{j=1}^m \kappa(g_i^{-1} x_j) \cdot v(x_j) \, q_j = \sum_{j=1}^m K_{ij} \cdot v(x_j) \, q_j 
\end{align}
with $K_{ij} = \kappa(g_i^{-1} x_j)$. In the case where $\kappa$ is compactly supported, $K_{ij}$ is a sparse matrix with the number of non-zero entries per row depending on the resolution of the grid $x_j$ and the support of $\kappa$. To obtain a learnable filter, $\kappa$ is parametrized as a linear combination of a chosen set of basis functions.

\subsection{DISCO convolutions in one dimension}
\label{app:disco_1d}

For the sake of simplicity, we discuss the simple one-dimensional case on $D = [0,1]$ with periodic boundary conditions. We note that this corresponds to the circle group (or the torus) $\mathbb{T}$. For any element $y \in D$, the corresponding group operation $T_y$ is the translation $T_y: D \to D$, $x \mapsto x \oplus y$, where we denote by $x \oplus y$ a modular shift such that the result remains in $D$.

Then, the DISCO convolution in one dimension, for Lebesgue square-integrable functions $v$ and $\kappa$ becomes
\begin{align}
    (\kappa \star v)(y) &= \int_{[0,1]} \kappa(T_y^{-1}x)v(x)\,\mathrm{d}x = \int_{[0,1]} \kappa(x-y) v(x) \, \mathrm{d}x \approx \sum_{j=1}^m \kappa(x_j-y) \, v(x_j) \, q_j,
\end{align}
for suitable quadrature points $D^h = \{x_j\}_{j=1}^m$ with corresponding quadrature weights $\{q_j\}_{j=1}^m$.

To parameterize the filter $\kappa$, we pick a finite support $[0, x_\text{cutoff}]$ with $L$ equidistant collocation points $\xi^{(\ell)}\in [0, x_\text{cutoff}]$ and corresponding hat functions. The $\ell$-th hat function is then defined as
\begin{equation}\label{eq:disco_basis_1d}
    \kappa^{(\ell)}(x) =
    \left\{\begin{array}{lr}
            \frac{x - \xi^{(\ell-1)}}{\xi^{(\ell)} - \xi^{(\ell-1)}} & \text{for  } \xi^{(\ell-1)} \leq x < \xi^{(\ell)}\\[0.4em]
            \frac{\xi^{(\ell+1)} - x}{\xi^{(\ell+1)} - \xi^{(\ell)}} & \text{for  } \xi^{(\ell)} \leq x < \xi^{(\ell+1)}\\[0.4em]
            0 & \text{else,}
            \end{array}\right.
\end{equation}
where $\xi^{(0)},\xi^{(L+1)} \in [0, x_\text{cutoff}]$ are suitable boundary points.
The resulting filter is obtained as a linear combination $\kappa = \sum^{L}_{\ell=1} \theta^{(\ell)}k^{(\ell)}$
with trainable parameters $\theta^{(\ell)}$. Plugging this into \eqref{eq:disco_convolution}, we obtain the trainable DISCO convolution
\begin{equation}
\label{eq:1d_disco}
    \sum_{j=1}^m \kappa(x_j-y_i) \, v(x_j) \, q_j = \sum^{L}_{\ell=1} \sum_{j=1}^m \theta^{(\ell)} \, \kappa^{(\ell)}(x_j-y_i) \, v(x_j) \, q_j = \sum^{L}_{\ell=1} \sum_{j=1}^m \theta^{(\ell)} K^{(\ell)}_{ij} \,  v(x_j) \, q_j,
\end{equation}
where $K^{(\ell)}_{ij}=\kappa^{(\ell)}(x_j-y_i)$ are the shifted filter functions. 

Let us now consider the special case of an equidistant grid $D^h$, i.e., $x_{j+1} - x_{j} = h$, and a trapezoidal quadrature rule $q_j = h$.
Let us further assume that the output points $y_i$ coincide with the grid points and that the collocation points $\xi^{(\ell)}$ are given as the first $L$ grid points. Then, due to the property of the hat functions, $K^{(\ell)}_{ij}$ can only contain either $0$ or $1$ and we obtain the circulant convolution matrices given by
\begin{equation*}
    K^{(1)} = (e_1, e_2, e_3, \dots, e_m), \hspace{1cm} K^{(2)} = (e_2, e_3, \dots, e_{m}, e_1), \hspace{1cm} K^{(3)} = (e_{3}, \dots, e_{m}, e_1, e_{2}), \quad \hdots
\end{equation*}
where $e_i\in \mathbb{R}^m$ is the $i$-th standard basis vector. For regular grids on planar geometries,  we can thus efficiently implement the DISCO convolution in~\eqref{eq:1d_disco} in terms of highly-optimized CUDA kernels based on common-place convolutional layers. This observation also generalizes to higher dimensions, where the same discrete kernel is obtained when the kernel is continuously shifted on the grid.

\subsection{DISCO convolutions on the sphere}
\label{app:disco_sphere}

For general group actions $g \in SO(3)$, the outcome of the group convolution \eqref{eq:group_conv} will be a function defined on $SO(3)$. This is due to $\mathbb{S}^2$ not being a group but rather a manifold on which $SO(3)$ acts. We can see this by fixing the north-pole $n=[0,0,1]^\top$ and applying any rotation $g \in SO(3)$ to it. This will trace out the whole sphere despite the north pole eliminating one of the Eulerian rotation angles. Therefore, to ensure that the result of the convolution is still a function defined on $\mathbb{S}^2$, we can simply restrict $g$ in \eqref{eq:group_conv} to rotations in $SO(3)/SO(2)$, which is isomorphic to $\mathbb{S}^2$. Formally, this can be achieved by fixing the first of the three Euler angles parameterizing $g$ to 0.

\begin{figure}[htb]
    \centering
    \begin{subfigure}{.19\linewidth}
        \centering
        \includegraphics[width=0.95\linewidth, trim={20px 20px 20px 20px}, clip]{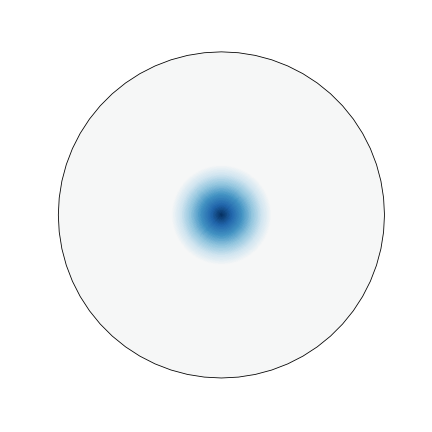}
        \caption{$\ell=1$}
    \end{subfigure}
    \begin{subfigure}{.19\linewidth}
        \centering
        \includegraphics[width=0.95\linewidth, trim={20px 20px 20px 20px}, clip]{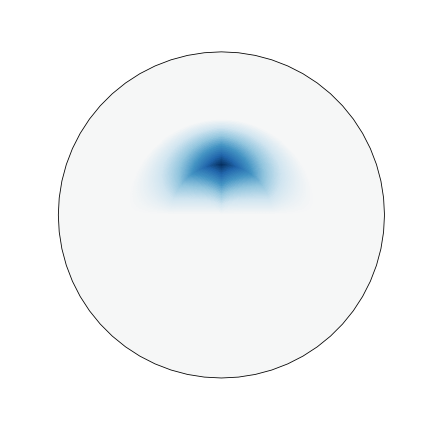}
        \caption{$\ell=2$}
    \end{subfigure}
    \begin{subfigure}{.19\linewidth}
        \centering
        \includegraphics[width=0.95\linewidth, trim={20px 20px 20px 20px}, clip]{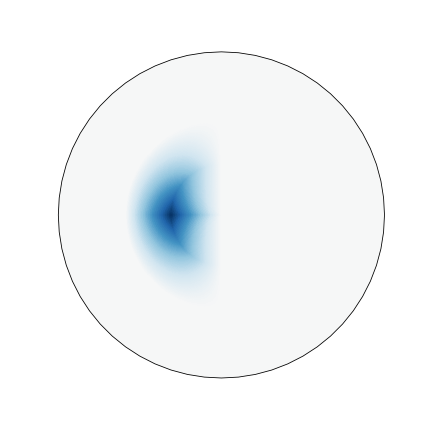}
        \caption{$\ell=3$}
    \end{subfigure}
    \begin{subfigure}{.19\linewidth}
        \centering
        \includegraphics[width=0.95\linewidth, trim={20px 20px 20px 20px}, clip]{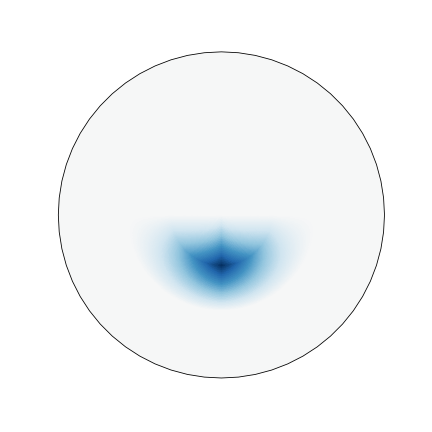}
        \caption{$\ell=4$}
    \end{subfigure}
    \begin{subfigure}{.19\linewidth}
        \centering
        \includegraphics[width=0.95\linewidth, trim={20px 20px 20px 20px}, clip]{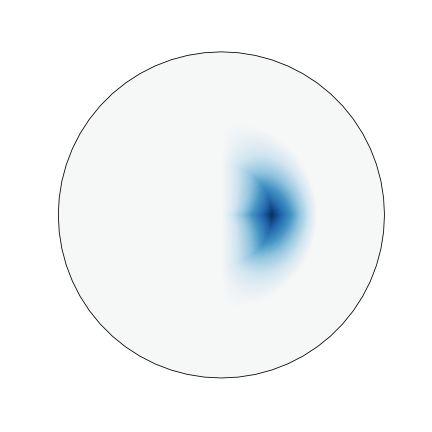}
        \caption{$\ell=5$}
    \end{subfigure}
    \caption{Radial, piecewise linear basis functions for the approximation of anisotropic filters on the sphere.}
    \label{fig:disco_basis}
    \vspace{-0.5em}
\end{figure}

As basis functions, we pick a set of piecewise linear basis functions as in \eqref{eq:disco_basis_1d}. To accommodate anisotropic kernels, collocation points are distributed in an equidistant manner along both radius and circumference. More precisely, the first collocation point lies at the center, and for each consecutive ring, a fixed amount of collocation points is distributed along the circumference. The resulting basis functions are illustrated in~\Cref{fig:disco_basis}, for a cutoff radius of $r_\text{cutoff} = 0.1 \pi$.

\section{Implementation details}
\label{sec:implementation_details}

A numerical comparison of our methods with baseline FNO and U-Net architectures can be found in~\Cref{tab:results}. In this section, we outline the implementation details for our numerical experiments.

For Darcy flow, Navier-Stokes, and the shallow water equations, training is conducted by minimizing the squared $L^2$-loss until convergence is reached. Our U-Net baseline is adapted from the model and code of the PDE Arena benchmark~\cite{gupta2022towards}.
For the FNOs~\cite{Li2020} and SFNOs~\cite{Bonev2023}, we use the implementation in the \texttt{neuraloperator} and \texttt{torch-harmonics} libraries. Moreover, for all experiments, GELU activation functions and the Adam optimizer are used.

For these three problems, we trained the FNO/SFNO-based models with varying widths and modes, while keeping the overall number of parameters approximately constant. We present the best results for each problem and macro-architecture in~\Cref{tab:results}. For the models with local layers, we found that a larger embedding dimension and fewer modes can often improve performance. We conjecture that the increased embedding dimension confers additional expressivity to the local kernels. This also suggests that local operators are an important inductive bias for these problems. When supplementing the FNO/SFNO blocks with additional branches, we also observed improved convergence by scaling the initial parameters or the output by $n^{-1/2}$, where $n$ is the number of branches. A detailed experimental setup is outlined for these three datasets in the following subsections. We also present results for the 2D diffusion-reaction equation and for super-resolution experiments on Darcy flow and the shallow water equations.

\subsection{2D Darcy flow equation}
\label{sec:exp_details_darcy}
\begin{figure}[htb]
    \centering
    \begin{subfigure}{.19\linewidth}
        \centering
        \includegraphics[width=0.95\linewidth, trim={20px 20px 20px 20px}, clip]{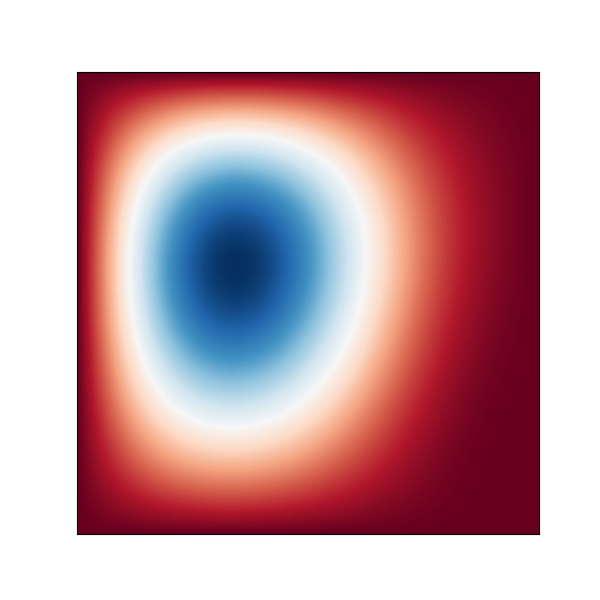}
        \caption{Input}
    \end{subfigure}
    \begin{subfigure}{.19\linewidth}
        \centering
        \includegraphics[width=0.95\linewidth, trim={20px 20px 20px 20px}, clip]{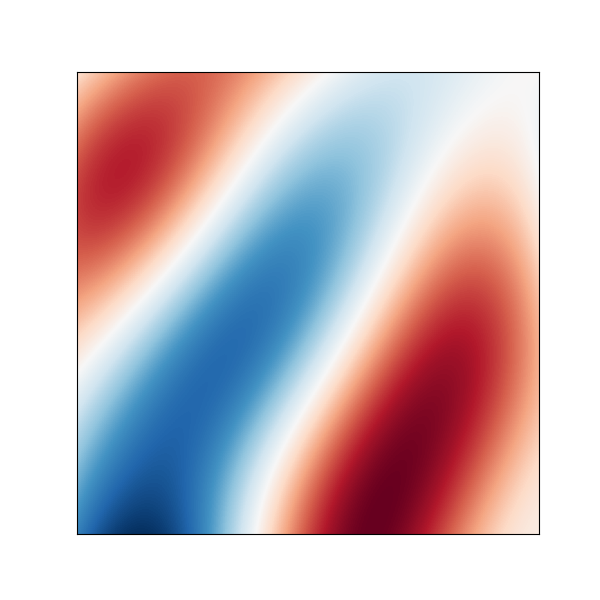}
        \caption{Ground truth}
    \end{subfigure}
    \begin{subfigure}{.19\linewidth}
        \centering
        \includegraphics[width=0.95\linewidth, trim={20px 20px 20px 20px}, clip]{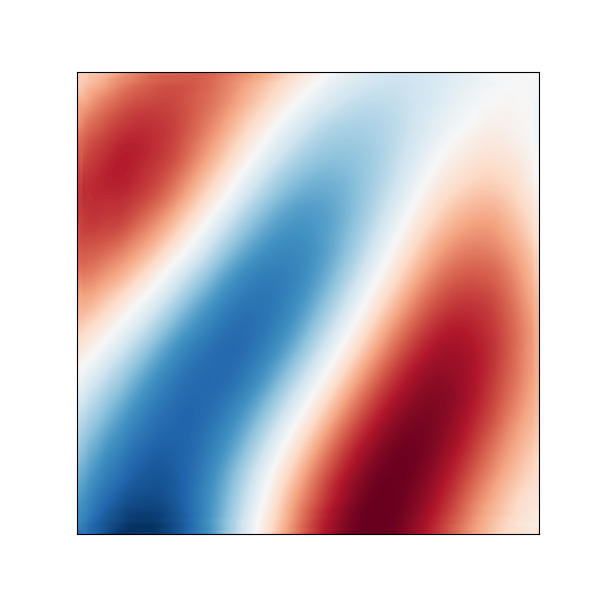}
        \caption{Ours}
    \end{subfigure}
    \begin{subfigure}{.19\linewidth}
        \centering
        \includegraphics[width=0.95\linewidth, trim={20px 20px 20px 20px}, clip]{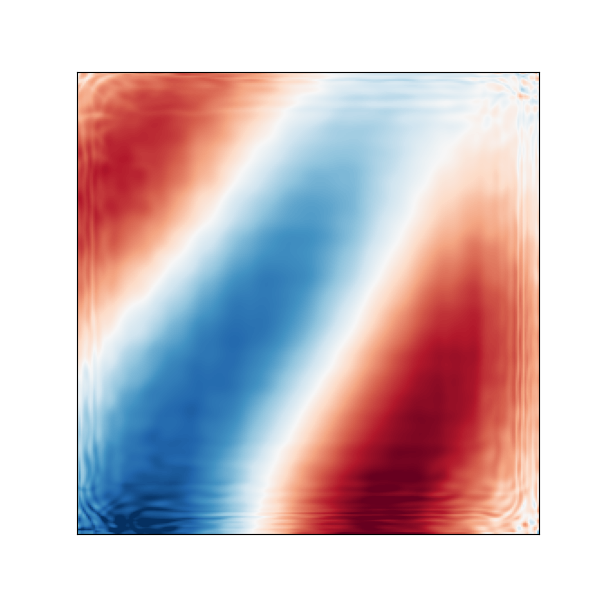}
        \caption{FNO}
    \end{subfigure}
    \begin{subfigure}{.19\linewidth}
        \centering
        \includegraphics[width=0.95\linewidth, trim={20px 20px 20px 20px}, clip]{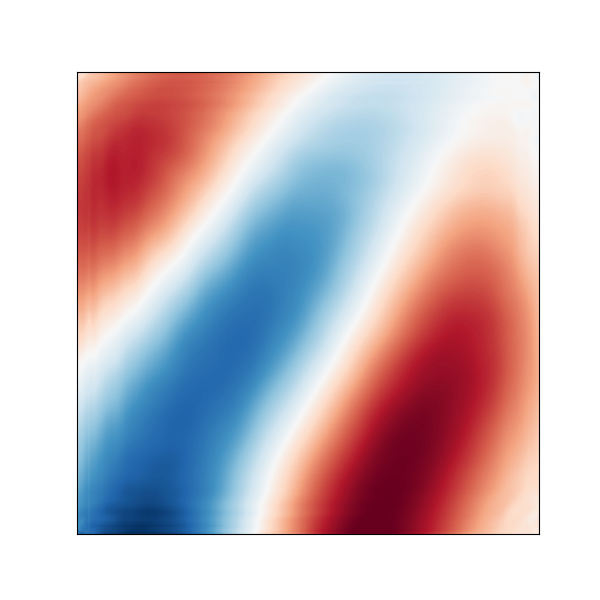}
        \caption{U-Net}
    \end{subfigure}
    \caption{Comparison of models from~\Cref{tab:results}: The outputs of our best-performing model, FNO, and U-Net on a randomly selected input pressure function from the Darcy flow problem. Edge artifacts are very prevalent in the FNO predictions, and they are less dominant in the U-Net predictions.}
    \label{fig:darcy_fig}
    \vspace{-0.5em}
\end{figure}

In the 2D Darcy flow setting, we generate our data as described in~\Cref{sec:darcy}. For the input functions, we consider random linear combinations of eigenfunctions of the Laplace operator with zero Dirichlet boundary conditions, i.e., 
\begin{equation*}
    u(x) = \sum_{i,j=1}^{20} \frac{c_{ij}}{\sqrt{(i\pi)^2 + (j\pi)^2}} \sin(i\pi x_1) \sin(j \pi x_2), \quad x\in D,
\end{equation*}
with i.i.d.\@ $c_{ij} \sim \mathcal{N}(0, 1/(i+j))$.
We train and test our models and baselines with data discretized onto a $256 \times 256$ regular grid. We use $10000$ training samples and $2000$ testing samples. As a baseline, we compare our proposed models with FNO~\cite{Li2020} and the U-Net architecture of \citet{gupta2022towards}. We note that this U-Net architecture is not agnostic to the discretization~\cite{Kovachki2021}, see also~\Cref{sec:standard_conv}. We compare these baselines against our proposed models: the architecture using convolutions with Fourier (i.e., FNO), differential, and integral kernels (\Cref{fig:layer}), as well as architectures using only FNO and differential kernels or only FNO with the proposed integral kernels. \Cref{fig:darcy_fig} compares the predictions of each model.

We choose all hyperparameters such that the overall number of parameters of all compared models is similar. The number of layers, number of Fourier modes, and embedding dimension are shown in~\Cref{tab:results}. Models using convolutions with differential and local integral kernels use these layers in parallel to the Fourier layers and pointwise skip connection on all layers. 
We use reflective padding for the convolutional operations in the differential and local integral kernels. 
For all relevant models, the local integral kernels use a radius cutoff of $0.007$ on $[-1,1]^2$, and they are parameterized by five radial, piecewise linear basis functions for the approximation of anisotropic filters (analogous to~\Cref{fig:disco_basis} on the plane). The differential kernels are parameterized as $3 \times 3$ convolutional kernels over the regular grid. For our U-Net baseline, we use $3 \times 3$ convolutional kernels with $2$ residual blocks for each resolution (two downsampling and two upsampling) and three layers within each block, with channel multipliers of $1,2,4$ for each layer within a block. 

Training is conducted by minimizing the squared $L^2$-loss for 70 epochs on a single NVIDIA P100 GPU, which is sufficient to achieve convergence on all models. We use a step learning rate decay scheduler, starting at $10^{-3}$ and halving every $10$ epochs. The results are shown in~\Cref{tab:results}.

\subsection{2D Navier-Stokes Equations}
\label{sec:exp_details_ns}
For the 2D Navier-Stokes equations (Kolmogorov flows), we use the same experimental setup and dataset as~\citet{mno_paper}, which sets $m = 4$ and $\textrm{Re}=5000$ in~\eqref{eq:navier_stokes} and uses a temporal discretization of $\tau = 1$ on a $128 \times 128$ regular grid. The initial conditions are sampled from a Gaussian measure as described in~\citet{mno_paper}, and the equation is solved with a pseudo-spectral solver. We compare the same five models as in the Darcy experiment.

As in the Darcy setting, we choose all hyperparameters such that the overall number of parameters for all the models is similar. The number of layers, number of Fourier modes, and embedding dimension are shown in~\Cref{tab:results}. Models using convolutions with differential and local integral kernels use these layers in parallel to the Fourier layers and pointwise skip connection on all layers.
We enforce periodic boundary conditions during padding for the convolutional operations in the differential and local integral kernels. 
For all relevant models, the local integral kernels use a radius cutoff of $0.05 \pi$ on the torus, and they are parameterized by five radial, piecewise linear basis functions for the approximation of anisotropic filters (analogous to~\Cref{fig:disco_basis} on the plane). The differential kernels are parameterized as $3 \times 3$ convolutional kernels over the regular grid. Our U-Net baseline is set up in the same way as in the Darcy experiment. 

Training was performed for 136 epochs on a single NVIDIA RTX 4090 GPU with an exponentially decaying learning rate, starting at $10^{-3}$ and halving every 33 epochs. The results are shown in~\Cref{tab:results}.

\subsection{Diffusion-Reaction equation}
\label{sec:exp_details_diffreac}
For the 2D Diffusion-Reaction equation, we use the same experimental setup and dataset as~\citet{takamoto2022pdebench}.   
The reference solution is computed using a finite-volume method in space and a fourth-order Runge-Kutta method in time. The dataset consists of $900$ training samples and $100$ validation samples discretized on a $128 \times 128$ regular grid with $100$ equidistant time-steps in the interval $[0,5]$ and Gaussian initial conditions. The task is to predict the state of the variables $(u,v)$ at the next time-step from the states at the previous $10$ steps. The errors are measured for the full autoregressive roll-out as in the implementation of PDEBench at \href{https://github.com/pdebench/PDEBench/}{github.com/pdebench/PDEBench}. 

We use the same experimental setup and implementation as~\citet {takamoto2022pdebench} and only adapt the learning rate, the number of modes, and the embedding dimension. We then add our proposed layers to (a subset of) the FNO blocks and experiment with both reflective and replicate padding. In particular, we parametrize the local integral kernels by five radial, piecewise linear basis functions and use $3 \times 3$ convolutional kernels for the differential kernels.  

We train on a single NVIDIA RTX 4090 GPU for 500 epochs with early stopping (using the same criteria as~\citet{takamoto2022pdebench}). Moreover, we use an exponentially decaying learning rate, starting at $10^{-4}$ and halving every 100 epochs. We present our best results in~\Cref{tab:diff_reac} and refer to~\citet{takamoto2022pdebench} for details on the metrics. We compare against the baselines by~\citet{takamoto2022pdebench} and note that we also improve their FNO baseline results. However, our local integral and differential kernels still provide a significant improvement with fewer parameters. Specifically, we reduced the modes from $24$ to $16$ and increased the embedding dimension from $20$ to $29$.

\subsection{Shallow water Equations}

For the shallow water equations, we use the dataset presented by \citet{Bonev2023}, which uses a Gaussian random field to generate initial conditions on an equiangular lat-lon-grid on the sphere at a resolution of $256\times 512$ and solves for $\varphi$, $u$ at a lead time of one hour. The target solution is computed using a spectral solver, which takes 24 Euler steps\footnote{Dataset and solver are taken from the \texttt{torch-harmonics} package at \href{https://github.com/NVIDIA/torch-harmonics}{github.com/NVIDIA/torch-harmonics}.}. Physical constants such as the sphere's radius or the Coriolis force are set to match those of Earth. The numerical solver uses $150$ explicit Euler steps to advance the solution $1$ hour in time. The right-hand side is discretized using the spectral basis provided by the Spherical Harmonics. For a detailed description of the dataset, we refer the reader to \citet{Bonev2023} and the corresponding implementation in the \texttt{torch-harmonics} package.

As a baseline for our experiments, we use the SFNO architecture as presented by \citet{Bonev2023}, where the embedding dimension is adjusted to $32$ to obtain a manageable parameter count. Moreover, we adapt the U-Net architecture by \citet{gupta2022towards} to the spherical domain by replacing all spatial (i.e., not the $1 \times 1$) convolutions with DISCO convolutions on the sphere. Therefore, the resulting architecture is a spherical U-Net similar to the architecture presented by \citet{Ocampo2022}. Moreover, due to the DISCO convolutions' discretization-agnostic nature, this architecture can be interpreted as a neural operator. Finally, we augment the SFNO architecture with local DISCO convolutions to obtain the proposed architecture; see~\Cref{sec:architecture}.

For all three architectures, hyperparameters were chosen to achieve roughly similar parameter counts. The learning rates for each architecture were determined with a quick parameter sweep, resulting in $3 \cdot 10^{-4}$ for the spherical U-Net and $3 \cdot 10^{-3}$ for both neural operators. As a learning rate scheduler, we use the policy of halving the learning rate upon a plateauing of the loss. Training was performed for $100$ epochs on a single NVIDIA RTX A6000 GPU, which was sufficient to achieve convergence on all considered models.

\subsection{Flow past a cylinder}
To demonstrate the capability of dealing with unstructured representations, we use the dataset provided by \citet{rahman2024pretraining} to train a model for predicting velocity and pressure fields for a Navier-Stokes problem in a channel with a suspended cylinder and attached membrane (see \Cref{fig:cylinder_example}). We use the dataset with a viscosity of $\mu=1.0$, which corresponds to a Reynolds number of $\text{Re}=2000$. To deal with the unstructured data, we employ a local integral convolution layer in both the encoder and decoder with a cutoff radius of $r_\text{cutoff} = 0.052$ and 5 piecewise-linear basis functions. This layer transforms the unstructured data to a regularly spaced grid of $48 \times 192$ in the internal representation. The processor part of the architecture then consists of 4 FNO blocks with local integral kernels, which have the same filter basis as the convolutions used in the encoder/decoder. The embedding dimension is fixed at 16 and all Fourier modes are kept in the internal representation. Overall, this leads to a parameter count of $9.6 \cdot 10^6$. The architecture is trained on a dataset of 250 samples for 50 epochs using the Adam optimizer and a learning rate of $2\cdot10^{-3}$. The results alongside baselines from \citet{rahman2024pretraining} are reported in \Cref{tab:cylinder}.

\subsection{Zero-shot super-resolution results}
\label{sec:superres}

\begin{figure}[t]
    \centering
    \begin{subfigure}{.24\linewidth}
        \centering
        \includegraphics[width=0.95\linewidth, trim={20px 20px 20px 20px}, clip]{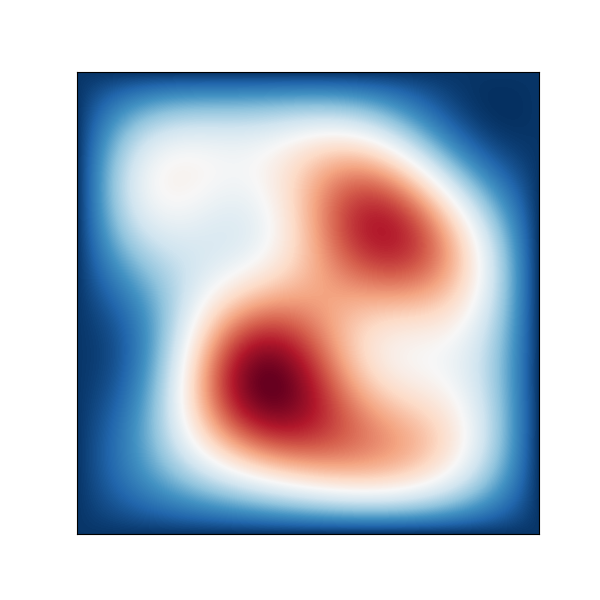}
        \caption{Input pressure}
    \end{subfigure}
    \begin{subfigure}{.24\linewidth}
        \centering
        \includegraphics[width=0.95\linewidth, trim={20px 20px 20px 20px}, clip]{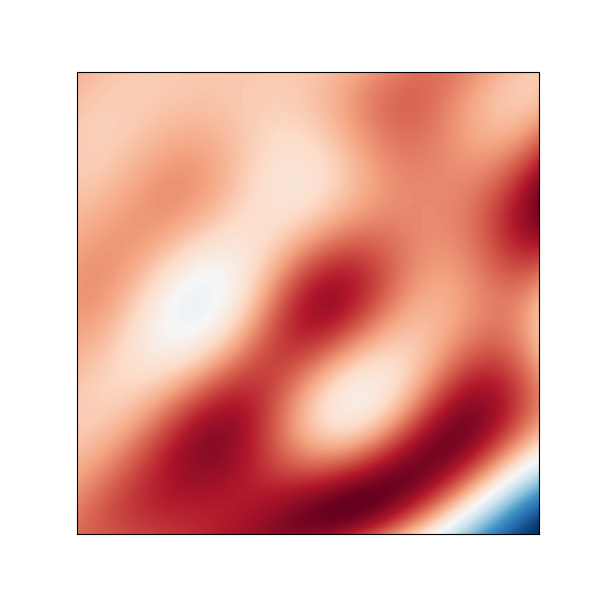}
        \caption{Ground truth forcing}
    \end{subfigure}
    \begin{subfigure}{.24\linewidth}
        \centering
        \includegraphics[width=0.95\linewidth, trim={20px 20px 20px 20px}, clip]{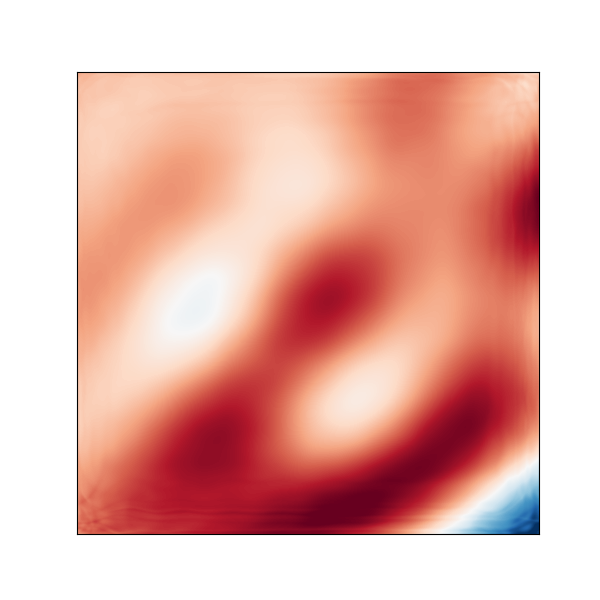}
        \caption{Predicted forcing}
    \end{subfigure}
    \\
    \begin{subfigure}{.24\linewidth}
        \centering
        \includegraphics[width=0.95\linewidth, trim={20px 20px 20px 20px}, clip]{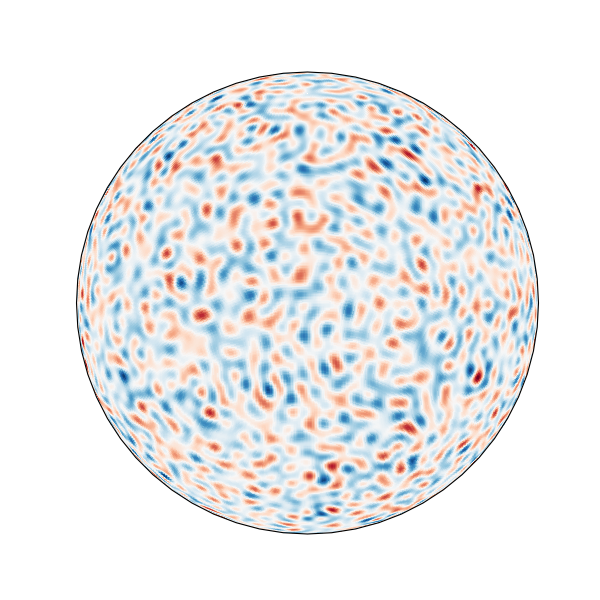}
        \caption{Initial geopotential height}
    \end{subfigure}
    \begin{subfigure}{.24\linewidth}
        \centering
        \includegraphics[width=0.95\linewidth, trim={20px 20px 20px 20px}, clip]{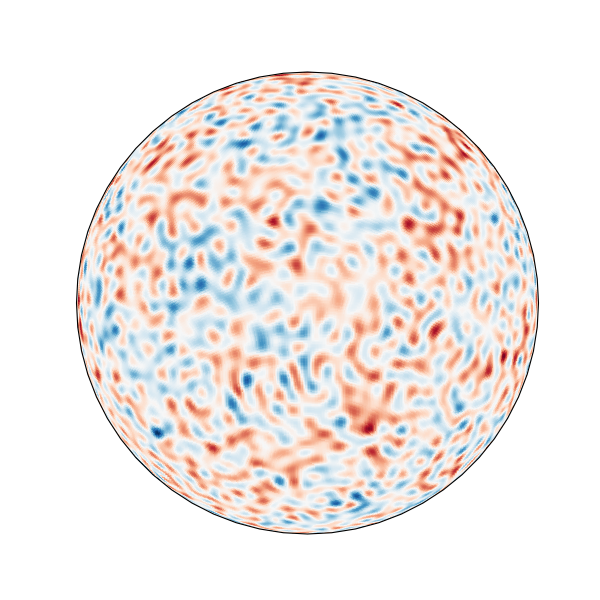}
        \caption{Ground truth at 1h}
    \end{subfigure}
    \begin{subfigure}{.24\linewidth}
        \centering
        \includegraphics[width=0.95\linewidth, trim={20px 20px 20px 20px}, clip]{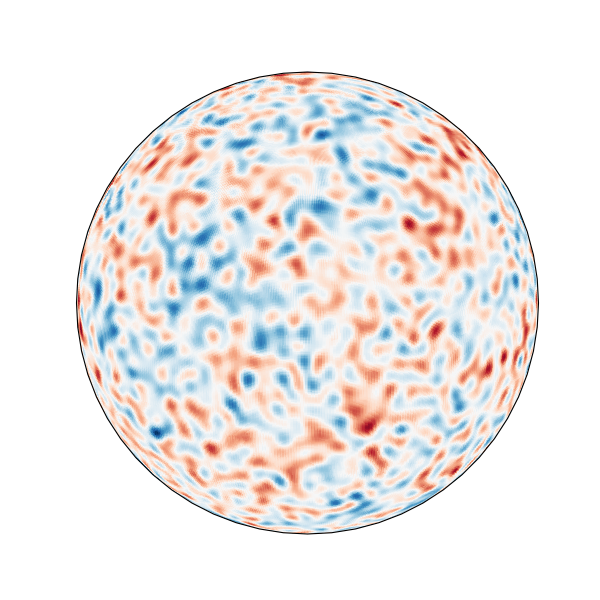}
        \caption{Prediction at 1h}
    \end{subfigure}
    \caption{Randomly selected super-resolution samples for the Darcy flow (top row) and shallow water (bottom row) problems.}
    \label{fig:superres}
    \vspace{-0.5em}
\end{figure}

\begin{table*}[tb!]
    \vskip 3pt
    \setlength\tabcolsep{1.9pt}
    \caption{Zero-shot super-resolution results for Darcy flow and the spherical shallow water problems. The validation error is reported in terms of relative $L^2$-error at various resolutions. Autoregressive rollouts are super-resolved in the sense that the rollout is performed at the high resolution.}
    \label{tab:superres}
    \vspace{-0.5em}
    \begin{center}
    \begin{small}
    \addtolength{\tabcolsep}{2pt}
    \begin{tabular}{lcccccccc}
    \toprule
    \multirow{2}{*}[-1pt]{Model} & \multicolumn{4}{c}{Parameters} & \multicolumn{4}{c}{Relative $L^2$-Error} \\
    \cmidrule(lr){2-5} \cmidrule(lr){6-9}
    & \multirow{2}{*}[-1pt]{\scriptsize{\# Layers}} & \multirow{2}{*}[-1pt]{\scriptsize{\# Modes}} & \multirow{2}{*}[-1pt]{\scriptsize{Embedding}} & \multirow{2}{*}[-1pt]{\scriptsize{\# Parameters}} & \multicolumn{4}{c}{resolution (relative to training resolution)} \\
    \cmidrule(lr){6-9}
    &  &   &  &   & $1/2 \times$ & $1 \times$ & $2 \times$ & $4 \times$ \\
    \midrule
    \multicolumn{9}{c}{Darcy Flow} \\
    \midrule
    FNO              & 4 & 20 & 41 & $2.715 \cdot 10^6$ & $1.475 \cdot 10^{-1}$ & $5.867 \cdot 10^{-2}$ & $8.646 \cdot 10^{-2}$ & $7.731 \cdot 10^{-2}$  \\
    \textbf{FNO + diff. (ours)} & 4 & 20 & 40 & $2.599 \cdot 10^6$ & $\mathbf{1.174 \cdot 10^{-1}}$ & $\mathbf{5.851 \cdot 10^{-2}}$ & $\mathbf{7.774 \cdot 10^{-2}}$ & $\mathbf{6.681 \cdot 10^{-2}}$ \\
    \midrule
    \multicolumn{9}{c}{Spherical Shallow Water Equations} \\
    \midrule
    Spherical U-Net & 17 & - & 32 & $1.639 \cdot 10^6$ & $5.586 \cdot 10^{-3}$ & $6.160 \cdot 10^{-4}$ & $3.386 \cdot 10^{-3}$ & $4.102 \cdot 10^{-3}$ \\
    SFNO            & 4 & 128 & 32 & $1.066 \cdot 10^6$ & $1.342 \cdot 10^{-3}$ & $9.220 \cdot 10^{-4}$ & $3.830 \cdot 10^{-3}$ & $4.419 \cdot 10^{-3}$ \\
    \textbf{SFNO + loc. int. (ours)} & 4 & 128 & 31 & $1.019 \cdot 10^6$ & $\mathbf{8.673 \cdot 10^{-4}}$ & $\mathbf{2.624 \cdot 10^{-4}}$ & $\mathbf{3.341 \cdot 10^{-3}}$ & $\mathbf{4.097 \cdot 10^{-3}}$ \\
    \bottomrule
    \end{tabular}
    \addtolength{\tabcolsep}{-2pt}
    \end{small}
    \end{center}
    \vspace{-0.5em}
\end{table*}

In this paper, we propose two methods to embed the inductive bias of locality into neural operator architectures. The key distinction between our proposed methods and CNN-based architectures is that our methods are agnostic to the discretization of the input function. In this section, we present and discuss the super-resolution capabilities of our proposed models. In particular, we focus on two examples: (1) the Darcy flow equation to showcase our differential layers on a regular Cartesian grid and (2) the shallow water equation to demonstrate super-resolution for our local integration on the sphere. The experimental setting that we consider is that of \emph{zero-shot super-resolution}. In particular, suppose that the model has been trained at a particular training resolution (or at multiple different resolutions). Given input at a higher resolution than the training resolution, the task of zero-shot super-resolution is to then predict the output function at this higher resolution and evaluate the resulting model error.

In the Darcy flow setting, we use the same setup and dataset as described in~\Cref{sec:darcy}. We train two models on data sampled at a $256 \times 256$ regular grid and evaluate on a $128 \times 128$ regular grid ($\frac{1}{2}$x resolution and $\frac{1}{4}$x the number of points), $512 \times 512$ grid (2x super-resolution), and $1024 \times 1024$ grid (4x super-resolution). For the shallow water equations, a similar approach is taken. We take the models in~\Cref{sec:swe} which were trained at a resolution of $256 \times 512$ and apply them to data generated at a resolution of $128 \times 256$, $512 \times 1024$, and $1024 \times 2048$.

\Cref{tab:superres} provides the details of the models we used for these experiments as well as the results of our super-resolution experiments with our proposed models. In the Darcy setting, we compare the baseline FNO to the FNO augmented with the differential kernel. On the sphere, we make use of the fact that the spherical U-Net presented in \cite{Ocampo2022} is already a neural operator due to its discretization-independence. As such, we can use it alongside the SFNO as a baseline for zero-shot super-resolution on the sphere.

As with all neural operator architectures, during training there is the possibility of overfitting to the training resolution. For instance, FNO may learn features in Fourier space that are intrinsically tied to the resolution of the input function. Similarly, it is possible that our proposed differential and integral convolutional operators will learn a function of the training discretization. This effect can be remedied by using high-resolution training data where the local details are fully resolved. This minimum required resolution of the training data is thus a function of the smoothness of the input function. For this reason, we decided to exclude the 2D Navier-Stokes problem from our super-resolution experiments, since training at a resolution that sufficiently resolves the local dynamics would be prohibitvely expensive.

In our experiments, we noticed that our differential layers tend to incur some discretization errors when trained on data that is not of sufficiently high resolution. If differential convolutions are present in consecutive layers in the model, this error can propagate quickly. As such, we found that the best-performing model on zero-shot super-resolution for the Darcy flow problem is a model with a differential convolution in only the first layer. Using fewer differential layers in our model reduces the expressivity, but we find that the super-resolution capabilities are still better than the baseline FNO.

Lastly, we would like to note that in our experiments, FNO had a larger error near the boundary in non-periodic problems, as the FFT used in FNO assumes periodic boundary conditions. We note that our proposed differential layer can help reduce the error at the boundary caused by FNO, but some of these effects may still be present (see~\Cref{fig:superres}).

\subsection{Computational efficiency of differential and local integral kernels}
On equidistant grids, our differential and local integral kernels can be implemented as standard convolutional kernels, which are heavily optimized for GPU performance. The computational complexity of our differential kernels is linear in the number of grid points since the size of the convolutional kernel remains constant regardless of resolution. In contrast, the size of the discretized local integral kernels does increase with input resolution. While this may be expensive for some high-resolution datasets, in our experiments, we found that the heavily-optimized convolutions in deep learning libraries can help reduce this computational burden. As a consequence, local integral kernels are more efficient than GNO~\citep{li2020neural} and other graph-based approaches on equidistant grids.

Moreover, the local integral kernels can also be applied to irregular grids, where they are implemented as sparse matrix multiplications. This tends to outperform GNOs~\citep{li2020neural}, which need to evaluate a neural network on the graph. The complexity of the local integral layer is linear with a constant that depends on the sparsity of the matrix. This, in turn, depends on the support of the local kernel; choosing a fixed support will scale quadratically with the resolution, whereas a support that matches the resolution will scale linearly in practice.
\end{document}